\documentclass[]{uiuc_paper}
\pdfoutput=1

\usepackage[toc,page,header]{appendix}
\usepackage{minitoc}
\usepackage{cleveref}
\usepackage{subcaption}
\usepackage{booktabs}
\usepackage{tabularx}

\usepackage{latexsym}
\usepackage{url}
\usepackage{amssymb}
\usepackage{amsmath}
\usepackage{amsfonts}
\usepackage[utf8]{inputenc}
\usepackage{booktabs}
\usepackage{pifont}
\usepackage{multirow}
\usepackage{makecell}
\usepackage{paralist}
\usepackage{xspace}
\usepackage{adjustbox}
\usepackage{needspace}
\usepackage{mathtools}
\usepackage{amsthm}
\usepackage{enumitem}
\usepackage{siunitx}
\usepackage{colortbl}
\usepackage{comment}
\usepackage{algorithm}
\usepackage{algorithmic}
\usepackage[framemethod=TikZ]{mdframed}

\mdfdefinestyle{findingbox}{%
  linecolor=blue!60,
  backgroundcolor=blue!5,
  roundcorner=10pt,
  innertopmargin=8pt,
  innerbottommargin=8pt,
  innerleftmargin=10pt,
  innerrightmargin=10pt,
  skipabove=8pt,
  skipbelow=8pt
}

\definecolor{lightblue}{RGB}{220,235,255}

\title{\(S^3\): Spectral Null-Space Swap Makes Reasoning Models Efficient}

\author[1,2]{Hongbo Ma$^*$}
\author[3,4]{Sansheng Cao$^*$}
\author[1]{Jiajun Fan}
\author[1]{Bangji Yang}
\author[1]{Ge Liu}

\affiliation[1]{University of Illinois Urbana-Champaign}
\affiliation[2]{Tsinghua University}
\affiliation[3]{Dartmouth College}
\affiliation[4]{Peking University}

\abstract{LLMs trained with Chain-of-thought excel in reasoning capability, but often come with excessive token cost. We find that the core of reasoning capacity lies in the Thinking model’s weight component within the null space of a projection defined by the corresponding Non-thinking model’s dominant singular directions, and removing the subspace component can largely improve reasoning efficiency without hurting the accuracy gained during thinking-mode post-training. Unlike existing efforts that mostly operate within the dominant subspace, we are the first to unveil the critical role of the null space and harness it for model optimization. Motivated by this finding, we propose \textbf{Spectral Null-Space Swap} ($S^3$), a training-free composition of paired Non-thinking and Thinking checkpoints. Our method keeps the Non-thinking model inside its own dominant subspace and takes the Thinking checkpoint outside it, improving reasoning efficiency while maintaining accuracy. We extensively evaluate $S^3$ on 2B–30B dense and mixture-of-experts (MoE) architectures spanning 28 evaluation environments across \textbf{mathematical, multimodal, and audio reasoning domains}. $S^3$ establishes new empirical Pareto Frontiers among training-free model composition strategies: across all settings, it reduces inference token overhead by an average of 27.4\% compared to full Thinking models while simultaneously improving overall task accuracy by 1.0 percentage point (e.g., yielding +8.3\% accuracy on HMMT25 alongside a 33.0\% token speedup). We further use attention entropy for explanation and find that the retained component produces more concentrated attention, and we use a simplified analytical model about optimization to demonstrate why null-space can effectively reduce attention entropy, thereby improving the efficiency of reasoning.}

\date{\today}
\correspondence{\email{geliu@illinois.edu}}

\begin{document}

% ======== Theorem environments ========
\newtheorem{theorem}{Theorem}[section]
\newtheorem{proposition}[theorem]{Proposition}
\newtheorem{lemma}[theorem]{Lemma}
\newtheorem{corollary}[theorem]{Corollary}
\newtheorem{definition}[theorem]{Definition}
\newtheorem{assumption}[theorem]{Assumption}
\newtheorem{remark}[theorem]{Remark}

\maketitle

\insert\footins{\noindent $^*$Equal contribution.}

\section{Introduction}
\label{intro}

Scaling language models enabled few-shot adaptation~\cite{brown2020language}, instruction tuning aligned them with user intent~\cite{ouyang2022training}, and chain-of-thought (CoT) prompting made reasoning explicit~\cite{wei2022chain}. Recent models extend deliberate reasoning across text, vision, and audio~\cite{yang2025qwen3technicalreport,bai2025qwen3vltechnicalreport,xu2025qwen3omnitechnicalreport}, while agents interleave reasoning with actions and observations~\cite{yao2023react,liu2024novel,xu2026ael,zhang2026heterogeneous,guo2026treebased,guo2026investigating} to handle extended interactions, multi-agent coordination, and sequential decision-making~\cite{huang2026robostream,ma2026thinkingblueprints,macadmorph,zhu2025pathology,zhu2026medeyes,ma2025styletailorpersonalizedfashionstyling}. However, overly lengthy CoT imposes heavy computational costs as workloads scale, making \emph{efficient reasoning} essential.

Post-training—especially reinforcement learning in thinking mode—improves reasoning through multi-domain fine-tuning, policy optimization, and reward modeling~\cite{liang2025boosting,liu2026socraticpo,xu2025epo,wang2026causalrm,pan2026uncertainty,snell2025scaling}, but inevitably inflates token consumption. Prior work explores explicit budgets~\cite{muennighoff2025s1}, adaptive stopping, token pruning, and weight interpolation to balance accuracy and efficiency. Nevertheless, a core question remains:
\emph{where in weight space does a Thinking model differ functionally from its Non-thinking counterpart, and which part of that difference can be removed without incurring full generation costs?}

\begin{figure}[t]
    \centering
    \includegraphics[width=\linewidth]{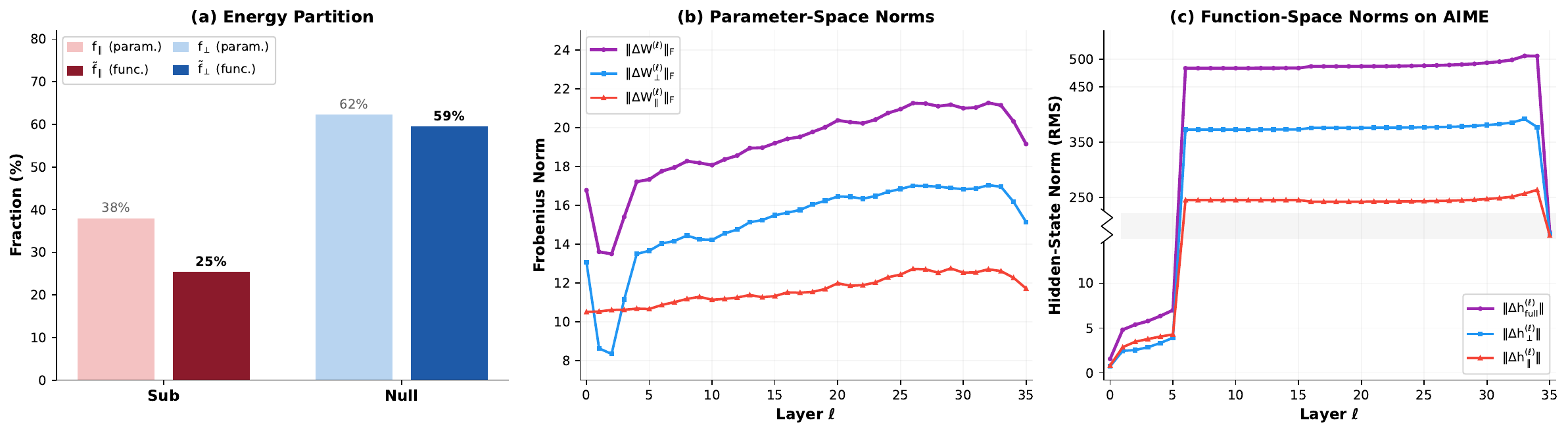}
    \caption{\textbf{This figure compares the parallel and perpendicular components of the thinking difference in parameter and function spaces.}
       All panels use the Qwen3-4B-Instruct and Qwen3-4B-Thinking. Parameter space denotes weight differences; function space denotes hidden-state offsets relative to Qwen3-4B-Instruct on AIME24/25. (a) Global average results. (b) Per-layer parameter-space results. (c) Per-layer function-space results on the same inputs.}
    \label{fig:insight}
\end{figure}

In spectral coordinates of the Non-thinking model, let $W_0$ and $W_t$ be paired Non-thinking and Thinking weights, with post-training difference $\Delta W=W_t-W_0$. We decompose $\Delta W$ into components aligned with and complementary to $W_0$'s singular subspace, comparing their Frobenius energies with induced hidden-state changes. Figure~\ref{fig:insight} reveals a structural asymmetry for Qwen3-4B on AIME24/25: the aligned component holds sizable parameter energy but induces weak functional shifts, whereas the complementary component drives most functional change despite lower energy.
Thus, weight changes along directions the Non-thinking model already occupies have little functional effect; function resides mostly in the complement. Unlike prior studies that primarily manipulate components within the dominant subspace, this finding underscores the untapped potential of the complementary space and motivates \emph{keeping Non-thinking in the former and importing Thinking in the latter}.

We instantiate this idea as \textbf{Spectral Null-Space Swap} ($S^3$). For each target matrix, the leading singular directions of $W_0$ define a protected subspace $\mathcal S_\rho$; the composed weight is $W_\rho=P_{\mathcal S_\rho}(W_0)+(I-P_{\mathcal S_\rho})(W_t)$. The protected region therefore keeps the Non-thinking model's structure, while its complement is taken from the Thinking model. Existing weight-space approaches average or interpolate checkpoints, resolve merging conflicts, or retain Base-aligned updates. $S^3$ instead makes an asymmetric, subspace-specific choice about \emph{which model supplies each region} without the additional training.

We evaluate paired 2B--30B dense and MoE models across text, vision--language, and audio tasks. Across diverse architectures and multimodal benchmarks, $S^3$ achieves substantial token efficiency while maintaining or improving accuracy relative to Thinking baselines. Crucially, several configurations establish empirical Pareto-optimal accuracy--token trade-offs among evaluated merging methods. For instance, Qwen3-4B-$S^3$ gains 8.3 percentage points in accuracy on HMMT25 alongside a 33.0\% token reduction; on MathVista, Qwen3-VL-4B-$S^3$ reaches 78.05\% accuracy with 31.9\% fewer tokens. The $\rho=1$ ablation isolates spectral components: its null-space model nearly matches Thinking accuracy with over a quarter fewer tokens, whereas the aligned-subspace model degrades toward Non-thinking.

Finally, we use attention entropy to examine why this decomposition changes reasoning behavior. Across evaluated datasets, the null-space model concentrates attention more than Non-thinking models, whereas adding only the aligned component diffuses attention, yielding the empirical ordering $H_{\mathrm{Null}}<H_{\mathrm{Base}}<H_{\mathrm{Sub}}$. Our simplified analytical model provides an explanation of when this ordering arises, offering a conditional prediction for small perturbation that matches observed attention patterns and shorter reasoning traces.

Our main contributions can be summarized as follows:
\begin{itemize}
      \item \textbf{Spectral finding.} We characterize the post-training thinking difference in both parameter and function spaces and identify a mismatch between their energy shares, showing that components along directions already occupied by Non-thinking models yield less functional impact, which instead largely resides in the complementary space.
    \item \textbf{Selective composition.} We introduce \textbf{Spectral Null-Space Swap} ($S^3$), a training-free composition of Non-thinking and Thinking checkpoints. It preserves Non-thinking's leading spectral directions and imports Thinking's complementary component. This transfer differs from whole-checkpoint interpolation and methods retaining base-aligned updates.
    \item \textbf{Accuracy--efficiency results.} We evaluate five paired model families spanning 2B--30B dense and MoE architectures across text, vision--language, and audio reasoning. Across settings, $S^3$ achieves multiple Pareto-Optimal operating points and uses 27.4\% fewer tokens on average than Thinking while improving mean accuracy by 1.0 percentage point.
    \item \textbf{Mechanistic explanation.} We introduce attention entropy to explain how $S^3$ alters model behaviors via spectral decomposition. Specifically, we observe the persistent ordering $H_{\mathrm{Null}}<H_{\mathrm{Base}}<H_{\mathrm{Sub}}$ across evaluated tasks and provide a simplified analytical model.
\end{itemize}
\section{Related Work}
\label{rw}

\subsection{Efficient Reasoning and Chain-of-Thought Compression}
\label{rw:efficient_reasoning}

Chain-of-thought (CoT) improves LLM reasoning but increases decoding cost. Existing approaches address this trade-off through adaptive computation and post-training, including adaptive stopping, token pruning, partial verification, process-reward learning, and reinforcement-learning-based optimization~\cite{huang2026does,li2026spvla,dong2026vvs,fan2026incentivizing,fan2025adaptive,yang2026batched,liang2025ropo,cai2025reinforcement,cai2026vi}.
In contrast, our method operates directly in weight space: it preserves the Non-thinking model in a protected spectral subspace and transfers only the complementary component from its corresponding Thinking model. This retains the Thinking model's reasoning capability while reducing CoT overhead, without additional training or decoding changes.

\subsection{Weight-Space Model Composition and Spectral Subspace Transfer}
\label{rw:weight_merge}

Weight-space merging combines fine-tuned models without retraining, evolving from checkpoint averaging and task arithmetic to interference reduction via sign resolution and delta sparsification~\cite{wortsman2022model,ilharco2023taskarithmetic,yadav2023ties,yu2023dare}. For efficient reasoning, direct Non-thinking--Thinking interpolation offers an accuracy--cost trade-off without extra training~\cite{wu2025revisiting}.
Spectral methods exploit parameter subspaces, including task-matrix analysis, LoRA-SVD alignment, shared/task-specific subspace separation, and base-aligned RL update decomposition~\cite{gargiulo2025tasksingular,stoica2025knots,marczak2025notask,zhang2026spectralrewiring}. 
Our approach instead preserves the Non-thinking model's leading spectral subspace while transferring only the complementary component from the Thinking model, using Non-thinking as an anchor and Thinking as a capability donor for selective reasoning transfer rather than multi-task fusion.

\subsection{Attention Entropy and Reasoning Dynamics}
\label{rw:attention_entropy}

Attention entropy characterizes attention concentration and has been linked to Transformer attention dynamics and geometry~\cite{ghader-monz-2017-attention,zhai2023stabilizing,ryu-lewis-2021-accounting}.  Recent work further relates attention entropy to reasoning dynamics and chain-of-thought trajectories~\cite{wen2025sparse,zhang2026cot,guan2026beware,shang2025united}. 
We observe consistent reductions in attention entropy across models and datasets, consistent with prior studies linking more concentrated attention patterns to effective contextual processing and reasoning, as well as recent work relating entropy dynamics to reasoning efficiency~\cite{wen2025sparse,zhang-etal-2025-attention,xiong-etal-2026-etr}. Beyond this empirical observation, we provide the first theoretical analysis connecting null-space projection in parameter space to attention entropy.
\section{Method: Spectral Null-Space Swap}
\label{method}

This section develops the method in three steps. Section~\ref{sec:insight} asks where a Thinking model differs from its Non-thinking counterpart in weight space, finding that functional differences lie mostly outside the Non-thinking model's occupied directions. Section~\ref{sec:swap} converts this finding into an operator that preserves the Non-thinking model's leading spectral directions while importing the remainder from the Thinking model, controlled by a single parameter. Section~\ref{sec:rho-one} examines full protection, where the operator transfers only expressive complements and forms a four-variant family differing solely in retained components—laying the groundwork for the interpretability analysis in Section~\ref{sec:interpretability}.

\subsection{Insight: Where the Thinking Difference Lives}
\label{sec:insight}

Let $W_0$ and $W_t$ denote the weights of a paired Non-thinking and Thinking checkpoints from the same model family. The two checkpoints always descend from the same pretrained model and differ in whether they produce an extended chain of thought before answering~\cite{yang2025qwen3technicalreport,bai2025qwen3vltechnicalreport}. All the thinking adds is the difference $\Delta W = W_t - W_0$, which is available from the two released checkpoints alone.

We measure $\Delta W$ in two ways. Its \emph{parameter-space} size is its
Frobenius energy; its \emph{function-space} size is the change it induces in the
network's computation. To separate the two, let $W_0 = U\Sigma V^\top$ be a
compact SVD and $P_{\mathcal S}(X) = U(U^\top X V)V^\top$ the orthogonal
projection onto the subspace $\mathcal S$ spanned by the singular directions of
$W_0$. Since $P_{\mathcal S}(W_0) = W_0$, this splits the difference into the part that lies where the anchor weight already has support and its complement---for the rectangular attention projections, exactly the null-space component:
\begin{equation}
\Delta W = \underbrace{P_{\mathcal S}(\Delta W)}_{\Delta W_{\parallel}} +
\underbrace{(I - P_{\mathcal S})(\Delta W)}_{\Delta W_{\perp}},
\qquad
\|\Delta W\|_F^2 = \|\Delta W_\parallel\|_F^2 + \|\Delta W_\perp\|_F^2 ,
\label{eq:split}
\end{equation}
with parameter-space share
$f_\parallel^{(\ell)} = \|\Delta W_\parallel^{(\ell)}\|_F^2 /
\|\Delta W^{(\ell)}\|_F^2$ and $f_\perp^{(\ell)} = 1 - f_\parallel^{(\ell)}$.

For the function-space counterpart, write $h_M^{(\ell)}(x) \in
\mathbb{R}^{T_x \times d}$ for the hidden states of model $M$ on input $x$ at
layer $\ell$, where $\ell = 0$ is the embedding output and $\ell = 1, \dots, L$
are the outputs of the transformer layers, and let
\begin{equation}
\Delta h_\ast^{(\ell)}(x) = h_{M_\ast}^{(\ell)}(x) - h_{W_0}^{(\ell)}(x),
\qquad
M_\ast \in \{W_0 + \Delta W_\parallel,\; W_0 + \Delta W_\perp,\;
W_0 + \Delta W\}
\label{eq:shift}
\end{equation}
be the shift that each variant induces relative to the anchor. Because the
shifts are read off actual forward passes, they capture the full nonlinear
effect of a weight change rather than a first-order Jacobian approximation.
Averaging the squared token-wise shifts over the $N = \sum_x T_x$ tokens of a
shared input set $X$ gives the functional norm and the corresponding share
\begin{equation}
\big\|\Delta h_\ast^{(\ell)}\big\|^2 = \frac{1}{N} \sum_{x \in X}
\sum_{t=1}^{T_x} \big\|\Delta h_\ast^{(\ell)}(x)_t\big\|_2^2,
\qquad
\tilde f_{\ast}^{(\ell)} = \frac{\|\Delta h_\ast^{(\ell)}\|^2}
{\|\Delta h_{\mathrm{full}}^{(\ell)}\|^2} .
\label{eq:func}
\end{equation}

Figure~\ref{fig:insight} compares the two decompositions. The split is stable across the middle and late layers, while the parameter-space and function-space measures consistently show that the component inside S accounts for much of the Frobenius energy but far less of the functional change, while the complementary component has similar shares under both measures. This suggests that a substantial portion of the difference lies along directions the anchor already occupies, yet contributes little to the model's actual computation. Thus, the functionally critical part of the thinking difference lies mainly outside the Non-thinking model’s spectral subspace, and the in-subspace component is largely redundant. This identifies which side the composed model should keep from the anchor and lays the foundation for model composition. Section~\ref{sec:swap} turns this observation into a concrete operator.

\begin{figure}[t]
  \centering
  \includegraphics[width=\linewidth]{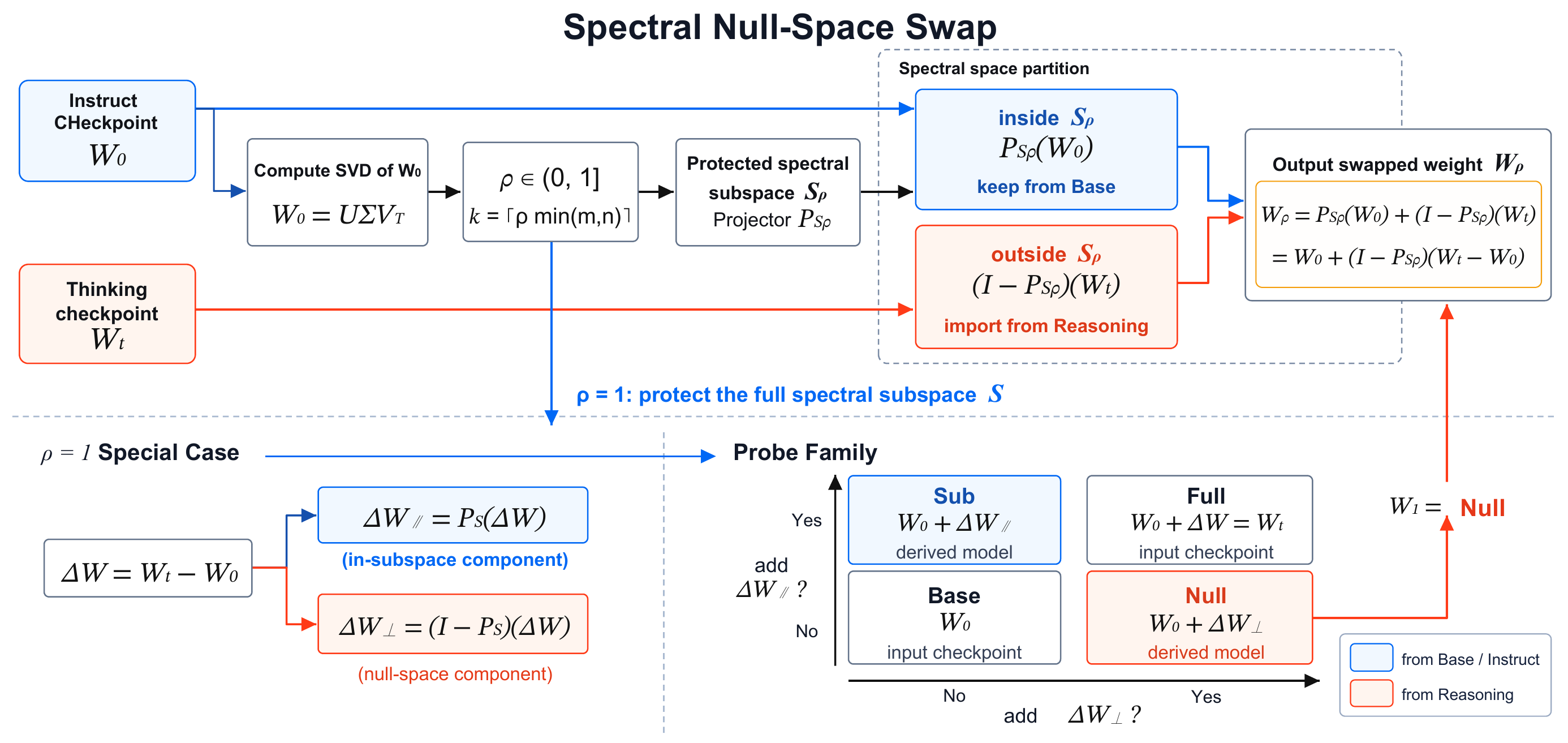}
  \caption{\textbf{Overview of our method: Spectral Null-Space Swap and the $\rho=1$ probe family.} The swapped weight preserves the protected spectral component of the Non-thinking model and imports the complementary component of the Thinking model. The four probes isolate the contributions of $\Delta W_{\parallel}$ and $\Delta W_{\perp}$.}
  \label{fig:overview}
\end{figure}

\subsection{The Spectral Null-Space Swap}
\label{sec:swap}

Section~\ref{sec:insight} splits the difference along the singular directions of
$W_0$. We now turn that split into an operator by letting only the leading
fraction of those directions play the role of $\mathcal S$. For
$\rho \in (0,1]$ let $U_k, V_k$ collect the leading
$k = \lceil \rho \min(m,n) \rceil$ singular vectors and
\begin{equation}
P_{\mathcal S_\rho}(X) = U_k (U_k^\top X V_k) V_k^\top,
\qquad
\mathcal S_\rho = \mathrm{span}(U_k) \otimes \mathrm{span}(V_k)
\subseteq \mathcal S .
\label{eq:proj-rho}
\end{equation}
As Figure~\ref{fig:overview} illustrates, the \textbf{Spectral Null-Space Swap} keeps the Non-thinking model inside $\mathcal S_\rho$ and
takes the Thinking model outside it,
\begin{equation}
W_\rho \;=\; P_{\mathcal S_\rho}(W_0) \;+\; (I - P_{\mathcal S_\rho})(W_t).
\label{eq:swap}
\end{equation}

Equation~\eqref{eq:swap} admits two equivalent interpretations. From a subspace perspective,
\begin{equation}
P_{\mathcal S_\rho}(W_\rho)=P_{\mathcal S_\rho}(W_0),\qquad
(I-P_{\mathcal S_\rho})(W_\rho)=(I-P_{\mathcal S_\rho})(W_t),
\label{eq:invariants}
\end{equation}
which shows that the two spectral regions are inherited from different sources: inside the protected subspace $\mathcal S_\rho$, the composed weight exactly matches the Non-thinking model, whereas on the orthogonal complement it exactly matches the Thinking model. Thus, $\mathcal S_\rho$ acts as the interface separating the structure preserved from $W_0$ from the component imported from $W_t$.

The same construction can be viewed from the perspective of transferring the difference. Rewriting \eqref{eq:swap} gives
\begin{equation}
W_\rho = W_0 + (I-P_{\mathcal S_\rho})(W_t-W_0),
\label{eq:transfer}
\end{equation}
so the composed model can be obtained by starting from $W_0$, decomposing the thinking difference $\Delta W=W_t-W_0$ into components inside and outside $\mathcal S_\rho$, discarding the former, and transferring only the latter. In this view, the operation does not interpolate the two checkpoints globally; instead, it selectively injects the part of the thinking difference that lies outside the protected spectral subspace of the Non-thinking model.

\subsection{Specialization at $\rho = 1$: The Probe Family}
\label{sec:rho-one}

Setting $\rho = 1$ in \eqref{eq:transfer} removes the only free parameter:
$k = \min(m,n)$, hence $U_k = U$ and $V_k = V$, and the swap returns
\begin{equation}
W_1 \;=\; W_0 + (I - P_{\mathcal S})(W_t - W_0) \;=\; W_0 + \Delta W_\perp ,
\label{eq:rho-one}
\end{equation}
a transfer of exactly the component of $\Delta W$ that $W_0$ cannot represent
(for the $m > n$ MLP matrices, the left-null-space component instead, so the
expression is well defined for every target weight). Keeping or discarding
each of the two components of $\Delta W$ then gives four variants,
\begin{equation}
\underbrace{W_0}_{\text{Base}},\qquad
\underbrace{W_0 + \Delta W_\parallel}_{\text{Sub}},\qquad
\underbrace{W_0 + \Delta W_\perp}_{\text{Null}},\qquad
\underbrace{W_0 + \Delta W}_{\text{Full}} ,
\label{eq:family}
\end{equation}
of which \textit{Base} and \textit{Full} are the released checkpoints and \textit{Sub} and \textit{Null} are derived from them. By \eqref{eq:rho-one} the \textbf{Spectral Null-Space Swap} is the Null entry, so the whole family is available from the two checkpoints alone, with no training, data, or roll-outs.

The construction is also convenient for interpretability: the four variants of \eqref{eq:family} differ only in which spectral component of a single difference is applied, and the decomposition is exact and local to each weight matrix, so an observed effect can be attributed to specific directions in a specific layer. Section~\ref{sec:interpretability} uses the family in this way.
\section{Experiments}
\label{experiments}
In this section, we first describe the evaluation settings and baselines across text, vision--language, and audio reasoning tasks in Section~\ref{exp:setup}. We then evaluate the cross-modal accuracy--efficiency trade-off of our method, study the effect of the protected-subspace ratio $\rho$, and isolate the respective contributions of the subspace and null-space components in Section~\ref{exp:main_result}, Section~\ref{exp:ratio-ablation} and Section~\ref{exp:ablation_space}. Additional selected-result visualizations and analyses, including Pareto-frontier plots and extended ratio sweeps, are provided in Appendix~\ref{app:cherry_pick}.

\begin{table*}[!ht]
\caption{\textbf{Main results across text, vision--language, and audio reasoning benchmarks.}
We report accuracy (\%) and average generated tokens per problem.
\textbf{Bold} indicates the best value in each column among TIES, MI-0.8, and S$^3$.
In each $\Delta$ row, accuracy is reported as an absolute change and token usage as a relative change with respect to the corresponding Thinking model.}
\label{tab:main-results}
\vskip 0.12in
\centering
\small
\resizebox{\textwidth}{!}{
\begin{tabular}{l|cc|cc|cc|cc}
\specialrule{1.2pt}{1pt}{1pt}
\multirow{2}{*}{\textbf{Large Language Model}}
& \multicolumn{2}{c|}{\textbf{AIME25}}
& \multicolumn{2}{c|}{\textbf{HMMT25}}
& \multicolumn{2}{c|}{\textbf{CMIMC25}}
& \multicolumn{2}{c}{\textbf{Olympiad-Bench}} \\
& Acc$\uparrow$ & Tok$\downarrow$
& Acc$\uparrow$ & Tok$\downarrow$
& Acc$\uparrow$ & Tok$\downarrow$
& Acc$\uparrow$ & Tok$\downarrow$ \\
\midrule
\multicolumn{9}{c}{\cellcolor{gray!15}\textbf{Qwen3-4B}} \\
\midrule
Qwen3-4B-Instruct
& 45.8 & 6,785 & 35.0 & 8,621 & 32.5 & 8,461 & 77.3 & 4,468 \\
Qwen3-4B-Thinking
& 71.7 & 20,704 & 46.7 & 24,636 & 58.1 & 24,415 & 83.8 & 14,086 \\
Qwen3-4B-TIES
& \textbf{73.3} & 16,377 & 50.0 & \textbf{16,049}
& 54.4 & 21,748 & \textbf{85.8} & 10,238 \\
Qwen3-4B-MI-0.8
& 71.7 & 17,270 & \textbf{58.3} & 19,729
& \textbf{56.3} & 22,902 & 85.5 & 11,460 \\
\textbf{Qwen3-4B-S$^3$-0.8}
& \textbf{73.3} & \textbf{15,004} & 55.0 & 16,496
& 55.0 & \textbf{21,266} & 83.8 & \textbf{9,745} \\
$\Delta$ \textit{vs.} Thinking
& \textcolor{teal}{$+$1.6} & \textcolor{teal}{$-$27.5\%}
& \textcolor{teal}{$+$8.3} & \textcolor{teal}{$-$33.0\%}
& \textcolor{red}{$-$3.1} & \textcolor{teal}{$-$12.9\%}
& \textcolor{teal}{$+$0.03} & \textcolor{teal}{$-$30.8\%} \\
\midrule
\multicolumn{9}{c}{\cellcolor{gray!15}\textbf{Qwen3-30B-A3B}} \\
\midrule
Qwen3-30B-A3B-Instruct
& 60.0 & 6,851 & 45.8 & 7,558 & 45.0 & 7,478 & 82.2 & 4,305 \\
Qwen3-30B-A3B-Thinking
& 84.2 & 17,579 & 68.3 & 21,093 & 64.4 & 21,979 & 85.8 & 12,359 \\
Qwen3-30B-A3B-TIES
& 72.5 & \textbf{10,472} & 52.5 & \textbf{15,555}
& 58.1 & 16,068 & 84.0 & \textbf{8,560} \\
Qwen3-30B-A3B-MI-0.8
& 76.7 & 15,648 & \textbf{65.0} & 19,432
& \textbf{68.1} & 19,632 & \textbf{86.9} & 10,899 \\
\textbf{Qwen3-30B-A3B-S$^3$-0.8}
& \textbf{80.8} & 12,536 & 62.5 & 16,424
& 66.3 & \textbf{15,988} & 88.4 & 9,238 \\
$\Delta$ \textit{vs.} Thinking
& \textcolor{red}{$-$3.37} & \textcolor{teal}{$-$28.7\%}
& \textcolor{red}{$-$5.8} & \textcolor{teal}{$-$22.1\%}
& \textcolor{teal}{$+$1.85} & \textcolor{teal}{$-$27.3\%}
& \textcolor{teal}{$+$2.55} & \textcolor{teal}{$-$25.3\%} \\
\specialrule{1.2pt}{1pt}{1pt}

\specialrule{1.2pt}{1pt}{1pt}
\textbf{Vision Language Model}
& \multicolumn{2}{c|}{\textbf{GSM8K}}
& \multicolumn{2}{c|}{\textbf{MMLU}}
& \multicolumn{2}{c|}{\textbf{MMMU}}
& \multicolumn{2}{c}{\textbf{MathVista-testmini}} \\
\midrule
\multicolumn{9}{c}{\cellcolor{gray!15}\textbf{Qwen3-VL-2B}} \\
\midrule
Qwen3-VL-2B-Instruct
& 87.88 & 918.5 & 65.68 & 484.0
& 46.00 & 6,665.8 & 57.73 & 2,235.4 \\
Qwen3-VL-2B-Thinking
& 92.75 & 2,295.4 & 80.15 & 3,228.0
& 55.50 & 8,075.4 & 66.90 & 3,623.8 \\
Qwen3-VL-2B-TIES
& \textbf{93.00} & 1,877.9 & 80.59 & 1,962.4
& \textbf{60.00} & 7,601.8 & 68.23 & 3,085.6 \\
Qwen3-VL-2B-MI-0.8
& 92.88 & 1,972.4 & 80.59 & 2,333.7
& 55.50 & 7,836.5 & 67.58 & 3,349.8 \\
\textbf{Qwen3-VL-2B-S$^3$-0.8}
& 91.62 & \textbf{1,550.1} & \textbf{80.92} & \textbf{1,754.8}
& 57.50 & \textbf{7,513.0} & \textbf{68.58} & \textbf{2,784.2} \\
$\Delta$ \textit{vs.} Thinking
& \textcolor{red}{$-$1.13} & \textcolor{teal}{$-$32.5\%}
& \textcolor{teal}{$+$0.77} & \textcolor{teal}{$-$45.6\%}
& \textcolor{teal}{$+$2.00} & \textcolor{teal}{$-$7.0\%}
& \textcolor{teal}{$+$1.68} & \textcolor{teal}{$-$23.2\%} \\
\midrule
\multicolumn{9}{c}{\cellcolor{gray!15}\textbf{Qwen3-VL-4B}} \\
\midrule
Qwen3-VL-4B-Instruct
& 96.37 & 266.1 & 78.62 & 57.8
& 50.25 & 597.4 & 73.35 & 1,104.9 \\
Qwen3-VL-4B-Thinking
& 96.00 & 1,403.0 & 90.02 & 1,823.9
& 80.00 & 5,152.7 & 76.02 & 2,843.9 \\
Qwen3-VL-4B-TIES
& \textbf{96.50} & 1,258.9 & 88.49 & 1,284.8
& 80.75 & 4,521.2 & \textbf{79.90} & 2,273.9 \\
Qwen3-VL-4B-MI-0.8
& 96.37 & 1,243.4 & \textbf{89.47} & 1,268.3
& 78.75 & 4,375.1 & 77.35 & 2,181.4 \\
\textbf{Qwen3-VL-4B-S$^3$-0.8}
& 96.00 & \textbf{1,139.6} & 88.49 & \textbf{1,167.3}
& \textbf{82.50} & \textbf{4,094.8} & 78.05 & \textbf{1,936.1} \\
$\Delta$ \textit{vs.} Thinking
& 0.00 & \textcolor{teal}{$-$18.8\%}
& \textcolor{red}{$-$1.53} & \textcolor{teal}{$-$36.0\%}
& \textcolor{teal}{$+$2.50} & \textcolor{teal}{$-$20.5\%}
& \textcolor{teal}{$+$2.03} & \textcolor{teal}{$-$31.9\%} \\
\specialrule{1.2pt}{1pt}{1pt}

\specialrule{1.2pt}{1pt}{1pt}
\textbf{Omni Language Model}
& \multicolumn{2}{c|}{\textbf{GSM8K}} 
& \multicolumn{2}{c|}{\textbf{MMLU}}
& \multicolumn{2}{c|}{\textbf{MMAR}}
& \multicolumn{2}{c}{\textbf{MMSU}}\\
\midrule
\multicolumn{9}{c}{\cellcolor{gray!15}\textbf{Qwen3-Omni-30B-A3B}} \\
\midrule
Qwen3-Omni-30B-A3B-Instruct
& 98.00 & 312 & 90.40 & 105
& 35.20 & 5,652 & 66.75 & 1,022 \\
Qwen3-Omni-30B-A3B-Thinking
& 97.60 & 936 & 92.80 & 1,166
& 63.30 & 982 & 70.33 & 622 \\
Qwen3-Omni-30B-A3B-TIES
& 97.20 & 860 & 92.90 & 693
& 64.97 & 799 & 66.47 & 530 \\
Qwen3-Omni-30B-A3B-MI-0.8
& \textbf{97.40} & 845 & \textbf{93.20} & 1,036
& \textbf{66.33} & 820 & 67.70 & 643 \\
\textbf{Qwen3-Omni-30B-A3B-S$^3$-0.8}
& 96.88 & \textbf{754.2} & 91.89 & \textbf{608.4}
& 64.70 & \textbf{740} & \textbf{68.18} & \textbf{526} \\
$\Delta$ \textit{vs.} Thinking
& \textcolor{red}{$-$0.72} & \textcolor{teal}{$-$19.4\%}
& \textcolor{red}{$-$0.91} & \textcolor{teal}{$-$47.8\%}
& \textcolor{teal}{$+$1.40} & \textcolor{teal}{$-$24.6\%}
& \textcolor{red}{$-$2.15} & \textcolor{teal}{$-$15.4\%} \\
\specialrule{1.2pt}{1pt}{1pt}
\end{tabular}
}
\vskip -0.08in
\end{table*}

\subsection{Experimental Setup}
\label{exp:setup}

\paragraph{Settings.}
We evaluate \textbf{Spectral Null-Space Swap} on paired Non-thinking and Thinking checkpoints from the Qwen3 family~\cite{yang2025qwen3technicalreport}. Our evaluation covers text, vision--language, and audio reasoning tasks. Specifically, we use Qwen3-4B, Qwen3-30B-A3B, Qwen3-VL-2B, Qwen3-VL-4B, Qwen3-Omni-30B-A3B as our backbone models~\cite{bai2025qwen3vltechnicalreport,xu2025qwen3omnitechnicalreport}. This design evaluates both cross-modal generalization and compatibility with dense and MoE architectures. We report \textbf{pass@1} for single-sample correctness and \textbf{avg@4} for average performance over four generations. Additional model configurations are provided in Appendix~\ref{app:configuration}.

\paragraph{Baselines.}
We compare our method with the original Non-thinking and Thinking checkpoints from each paired model family. We further compare against MI-0.8, a direct Instruct--Thinking weight-interpolation baseline using the $0.8$ interpolation setting~\cite{wu2025revisiting}, and TIES-Merging, which resolves parameter conflicts through trimming and sign election~\cite{yadav2023ties}. All methods operate on the same paired checkpoints and are evaluated using identical prompts and decoding configurations.

\subsection{Main Results: Cross-Modal Accuracy--Efficiency Trade-offs}
\label{exp:main_result}

% Table~\ref{tab:main-results} shows that our method consistently provides a favorable accuracy--efficiency trade-off across text, vision--language, and audio reasoning. Compared with the Instruct checkpoints, our method substantially improves reasoning performance while retaining markedly shorter outputs than the corresponding Thinking models. On text reasoning, our method recovers most of the reasoning gains of Thinking and exceeds it on several benchmarks, while reducing chain-of-thought length. Across vision--language and audio tasks, our method also frequently achieves the lowest token usage among weight-merging baselines, while maintaining competitive accuracy. These results indicate that \textbf{Spectral Null-Space Swap} transfers useful reasoning behavior without inheriting the full decoding overhead of Thinking models. Compared with the TIES and MI-0.8 baselines, our method exhibits a more favorable Pareto frontier in the accuracy--token trade-off. We provide results on additional benchmarks, together with a detailed Pareto-frontier analysis of the accuracy--token trade-off in Appendix~\ref{app:main_results}.

Table~\ref{tab:main-results} shows that our method consistently provides a favorable accuracy--efficiency trade-off across text, vision--language, and audio reasoning. Compared with Non-thinking models, it substantially improves reasoning while maintaining markedly shorter outputs than Thinking models. On text tasks, our method recovers most Thinking gains—and exceeds them on several benchmarks—while shortening chains of thought. Across vision--language and audio domains, it frequently achieves the lowest token usage among merging baselines while preserving competitive accuracy. Thus, \textbf{Spectral Null-Space Swap} transfers reasoning capability without the full decoding overhead. Compared with TIES and MI-0.8, our method exhibits a superior Pareto frontier; additional benchmarks and detailed trade-off analyses appear in Appendix~\ref{app:main_results}.

\subsection{Ablation on the Protected-Subspace Ratio $\rho$}
\label{exp:ratio-ablation}

We study the effect of the protected-subspace ratio $\rho$, which controls the fraction of leading singular directions of the Non-thinking checkpoint retained during spectral subspace swap. Table~\ref{tab:ratio-sweep-4b} reports the results of Qwen3-4B-S$^3$ across $\rho\in[0.5,1.0]$.

Overall, decreasing $\rho$ transfers a larger component from the Thinking model and generally increases reasoning capability, but also leads to longer generations. This trend is particularly clear on AIME25, where smaller ratios improve accuracy while increasing token usage. In contrast, larger ratios yield the shortest outputs on several benchmarks, reflecting stronger preservation of the efficient Non-thinking model behavior. We use $\rho=0.8$ as the default configuration because it offers a balanced trade-off, achieving strong reasoning performance while avoiding the substantially longer trajectories observed at smaller ratios. Additional ablation results are shown in Appendix~\ref{app:ratio-sweep}.

\begin{table*}[t]
\centering
\small
\caption{Effect of the protected-subspace ratio in Qwen3-4B-S$^3$ on the four benchmarks reported in the main table. We report accuracy (\%) and average generated tokens per problem. \textbf{Bold} indicates the best value in each column.}
\label{tab:ratio-sweep-4b}
\resizebox{\textwidth}{!}{
\begin{tabular}{c|cc|cc|cc|cc}
\toprule
\textbf{Model}
& \multicolumn{2}{c|}{AIME25}
& \multicolumn{2}{c|}{HMMT25}
& \multicolumn{2}{c|}{CMIMC25}
& \multicolumn{2}{c}{Olympiad-Bench} \\
\midrule
Qwen3-4B-S$^3$-1.0 & 69.2 & 15,129 & 45.8 & \textbf{15,920}
& 50.6 & \textbf{17,863} & 83.0 & \textbf{9,553} \\
Qwen3-4B-S$^3$-0.9 & 71.7 & 15,843 & 49.2 & 16,733
& 53.1 & 20,567 & 83.5 & 9,607 \\
Qwen3-4B-S$^3$-0.8 & 73.3 & \textbf{15,004} & \textbf{55.0} & 16,496
& \textbf{55.0} & 21,266 & 83.8 & 9,745 \\
Qwen3-4B-S$^3$-0.7 & 75.8 & 18,082 & 45.0 & 22,377
& 48.1 & 22,186 & \textbf{84.0} & 10,265 \\
Qwen3-4B-S$^3$-0.6 & 78.3 & 18,542 & 49.2 & 22,481
& 50.6 & 22,798 & 83.8 & 11,607 \\
Qwen3-4B-S$^3$-0.5 & \textbf{79.2} & 19,264 & 54.2 & 23,361
& 51.3 & 23,743 & 83.6 & 12,035 \\
\bottomrule
\end{tabular}
}
\end{table*}

\subsection{Ablation on Subspace and Null-Space Components}
\label{exp:ablation_space}
Although the method in Section~\ref{method} performs well across benchmarks, we must rule out that the subspace component alone would suffice. We therefore compare the \textit{Sub} and \textit{Null} models against the \textit{Base} and \textit{Full} models on math benchmarks. Table~\ref{tab:ablation} shows that the \textit{Sub} model stays close to the \textit{Base} model, while the \textit{Null} model matches the \textit{Full} model. Therefore, the reasoning gain is concentrated in the null-space component.

\begin{table}[t]
\centering
\small
\caption{\textbf{Accuracy and token length across five math benchmarks generated by the probe family.} The \textit{Sub} model is almost identical to the \textit{Base} model. The \textit{Null} model matches the \textit{Full} model's average accuracy while generating fewer tokens.}
\begin{tabular}{lccccccc}
\toprule
Model & AIME'24 & AIME'25 & AMC'23 & CMIMC'25 & HMMT'25 & Mean Acc & Mean len \\
\midrule
Base & 61.7 & 50.4 & 94.7 & 30.6 & 33.3 & 54.1 & 8{,}280 \\
Sub & 69.2 & 56.2 & 95.9 & 36.2 & 35.4 & 58.6 & 8{,}891 \\
Null & 77.1 & 74.6 & 99.1 & 51.2 & 49.2 & 70.2 & 15{,}071 \\
Full & 75.4 & 70.0 & 99.1 & 53.4 & 53.3 & 70.3 & 20{,}269 \\
\bottomrule
\end{tabular}
\label{tab:ablation}
\end{table}
\section{Mechanistic Analysis}
\label{sec:interpretability}

% To understand why the null-space model achieves comparable task performance with higher reasoning efficiency, we investigate the mechanism from the perspective of attention entropy.
To explain why the null-space model achieves comparable task performance with improved reasoning efficiency, we introduce attention entropy as an analytical measure and investigate its underlying mechanism.

% \subsection{Attention Entropy}
\subsection{Preliminary: Attention Entropy and Reasoning Efficiency}
\label{sec:attention-entropy}

For a query position $t$ with attention distribution $p_t$ over preceding tokens, the attention entropy is defined as~\cite{ryu-lewis-2021-accounting, zhai2023stabilizing} 
$$H_t = -\sum_j p_t(j)\log p_t(j)$$
Lower entropy indicates more concentrated attention. Following prior work linking late layers to reasoning~\cite{chen2026think,li2026stop}, we measure entropy at layers 32-35 in Appendix~\ref{sec:layer-selection}. As shown in Figure~\ref{fig:entropy-analysis}, we reveal a consistent ordering across all datasets tested, which also holds at the per-problem level:
\begin{align}
    H_{\mathrm{Null}} < H_{\mathrm{Base}} < H_{\mathrm{Sub}}.
\end{align}

Attention entropy provides a useful lens for interpreting the efficiency differences among model variants. Prior work~\cite{shang2025united} has viewed attention entropy as a proxy for cognitive load in LLMs, where diffuse attention imposes higher extraneous load, requiring longer CoT to reach the same accuracy~\cite{xu2026unveiling,wu2026when}. This is consistent with the subspace model exhibiting the highest entropy together with the longest reasoning traces and the highest hallucination rate on SimpleQA in Appendix~\ref{sec:hallucination}.

\begin{figure}[t]
    \centering
    \includegraphics[width=\textwidth]{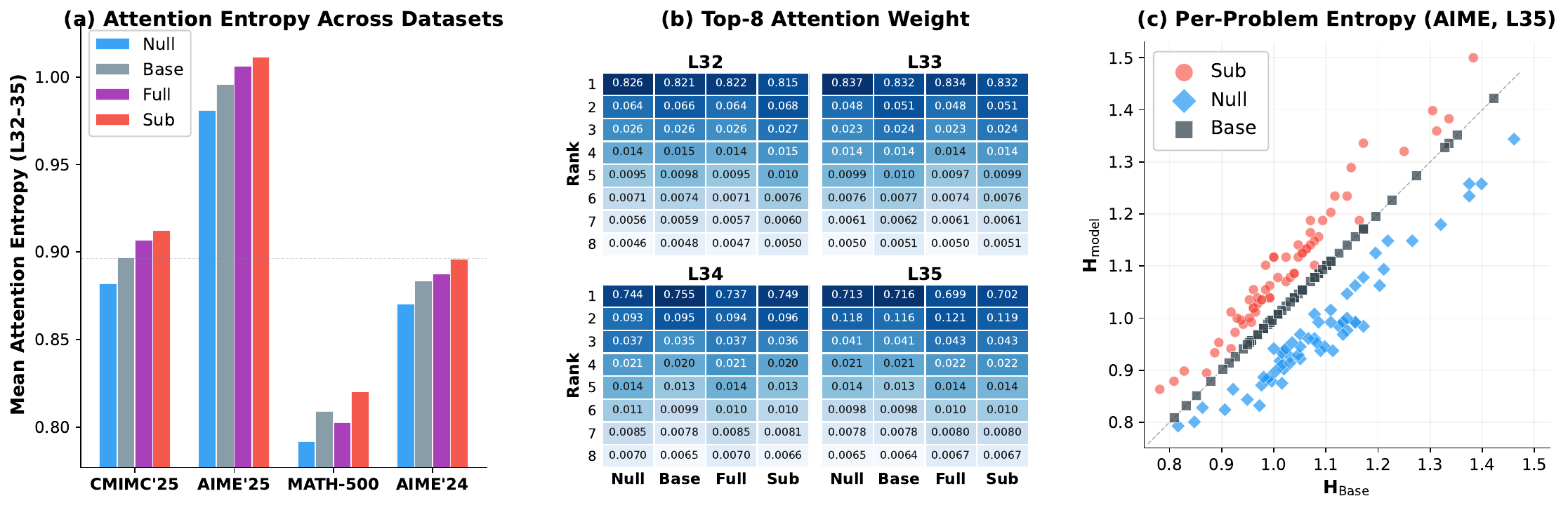}
    \caption{\textbf{Attention entropy analysis across four model variants.}
    (a)~Mean attention entropy averaged over layers 32-35: Null consistently achieves the lowest entropy while Sub exhibits the highest; additional results are shown in Appendix~\ref{sec:full-datasets}.
    (b)~Top-8 attention weights at layers 32-35 show that Null concentrates attention.
    (c)~Per-problem entropy scatter on AIME at layer 35: the ordering $H_{\text{Null}} < H_{\text{Base}} < H_{\text{Sub}}$ holds for all the individual problems.}
    \label{fig:entropy-analysis}
\end{figure}

% In conclusion, null-space yields more concentrated attention and better reasoning, raising the question of why it systematically reduces attention entropy, which we analyze in the next section.

In conclusion, null-space exhibits more concentrated attention and improved reasoning efficiency. We analyze the underlying mechanism that leads to its systematically lower attention entropy from a simplified theoretical perspective in Section~\ref{mec:theoretical}.

\subsection{A Simplified Analytical Model for the Subspace Entropy Ordering}
\label{mec:theoretical}
% The entropy ordering in Section~\ref{sec:attention-entropy} raises a natural question: why does the in-subspace component increase entropy while the null-space component reduces it? Our analysis builds on the observation that pretraining has already driven attention entropy to near-stationarity in Appendix~\ref{sec:saturation}, motivating a local-optimality assumption within the active subspace.
Given that gradient descent drives attention entropy to near-stationarity in Appendix~\ref{sec:saturation}, we adopt a heuristic local-optimality assumption within the subspace. This provides an intuitive explanation for why the in-subspace component tends to increase entropy while the null-space component reduces it.

\begin{assumption}[Local optimality in the active subspace]\label{ass:local-opt}
Consider the query projection $W_Q$ of a single attention head, and let $V = \mathrm{row}(W_{Q,0})$ denote the subspace (i.e. the row space) of the pretrained matrix. We assume that mean attention entropy is locally minimized within $V$: for any perturbation $\Delta W_\parallel \in V$,
\begin{align}
    \nabla_V H(W_{Q,0}) = 0, \qquad \phi_\parallel''(0) = \frac{d^2}{dt^2} H(W_{Q,0} + t\,\Delta W_\parallel)\Big|_{t=0} \geq 0.
\end{align}
\end{assumption}
While an idealization, this assumption captures the qualitative asymmetry between the active and null directions: the subspace has been directly shaped by the training process, whereas the orthogonal null space remains largely unconstrained by it. The entropy landscape is therefore expected to be locally stationary in the former while retaining nontrivial first-order variation in the latter.

\begin{proposition}[First-order null-space effect]\label{thm:first-order}
Let $G = \nabla_{W_Q} H(W_{Q,0})$ be the entropy gradient at the pretrained model. Under Assumption~\ref{ass:local-opt},
\begin{align}
    H(W_{Q,0} + \Delta W) - H(W_{Q,0}) = \langle \Delta W_\perp,\, G_\perp \rangle + O(|\Delta W|^2).
\end{align}
\end{proposition}

\begin{proposition}[Entropy increase under subspace perturbations]\label{thm:sub-increase}
Under Assumption~\ref{ass:local-opt},
\begin{align}
    H(W_{Q,0} + \Delta W_\parallel) - H(W_{Q,0}) = \frac{1}{2}\,\phi_\parallel''(0) + O(|\Delta W_\parallel|^3) \geq 0.
\end{align}
\end{proposition}

Proposition~\ref{thm:first-order} shows that only the null-space component $\Delta W_\perp$ contributes at first order, while the subspace component has a vanishing gradient within $V$. Proposition~\ref{thm:sub-increase} further implies that perturbations along the active subspace cannot decrease entropy locally, explaining why the Sub model consistently exhibits higher attention entropy than Base.

\begin{corollary}[Entropy ordering]\label{cor:ordering}
Under the same local approximation, given our empirical observation that $\langle \Delta W_\perp,\, G_\perp \rangle < 0$ (Appedix~\ref{app-subin}), it follows that for sufficiently small perturbations,
\begin{align}
   H_{\mathrm{Null}} \leq H_{\mathrm{Base}} \leq H_{\mathrm{Sub}}. 
\end{align}
Within this simplified model, the null-space projection isolates the entropy-reducing component, the subspace projection isolates the entropy-increasing component, and \textit{Full} interpolates between the two.
\end{corollary}

% Detailed proofs are given in Appendix~\ref{sec:proofs}.
Detailed proofs of the above propositions and corollary are given in Appendix~\ref{sec:proofs}.
\section{Conclusion}

% In this work, we investigate the functional distribution of difference in Thinking model and Non-thinking model in weight space 
In this work, we investigate where, in weight space, a Thinking model differs functionally from its corresponding Non-thinking model
and propose \textbf{Spectral Null-Space Swap ($S^3$)}, a training-free method for efficient reasoning. We identify a structural asymmetry between parameter energy and functional impact: the component of the difference aligned with the Non-thinking model's dominant singular subspace contains substantial parameter energy but induces weak functional changes, while the complementary null-space component drives the core functional shifts. By preserving the Non-thinking model inside its primary spectral subspace and importing only the Thinking model's complementary components, $S^3$ reduces reasoning token overhead without sacrificing performance. Across 2B--30B dense and Mixture-of-Experts (MoE) models and text, vision-language, and audio reasoning benchmarks, $S^3$ consistently improves the accuracy--efficiency trade-off. Across more than 25 evaluation settings, it reduces generated tokens by approximately 30\% while maintaining or improving accuracy over Thinking baselines. Mechanistically, null-space projection produces lower attention entropy and more concentrated attention, which we explain through local stationarity. Overall, $S^3$ provides both a geometric view of post-training weight difference and a simple training-free approach to efficient reasoning.

\bibliographystyle{unsrt}
\bibliography{main}

\clearpage
\appendix
\section{Mechanistic Analysis Details}
\label{mech_ana}
\subsection{Layer Selection}
\label{sec:layer-selection}

Prior work localizes reasoning relevant computation to the late layers of transformer language models and the reasoning state that governs when to stop can be read out from late-layer attention states. Attention entropy in the final layers therefore reflects the reasoning-critical computation that distinguishes the model variants. Accordingly, we measure attention entropy at layers 32--35, the final four layers of Qwen3-4B.

\subsection{Full Dataset Attention Entropy Results}
\label{sec:full-datasets}

\begin{figure}[H]
    \centering
    \includegraphics[width=\textwidth]{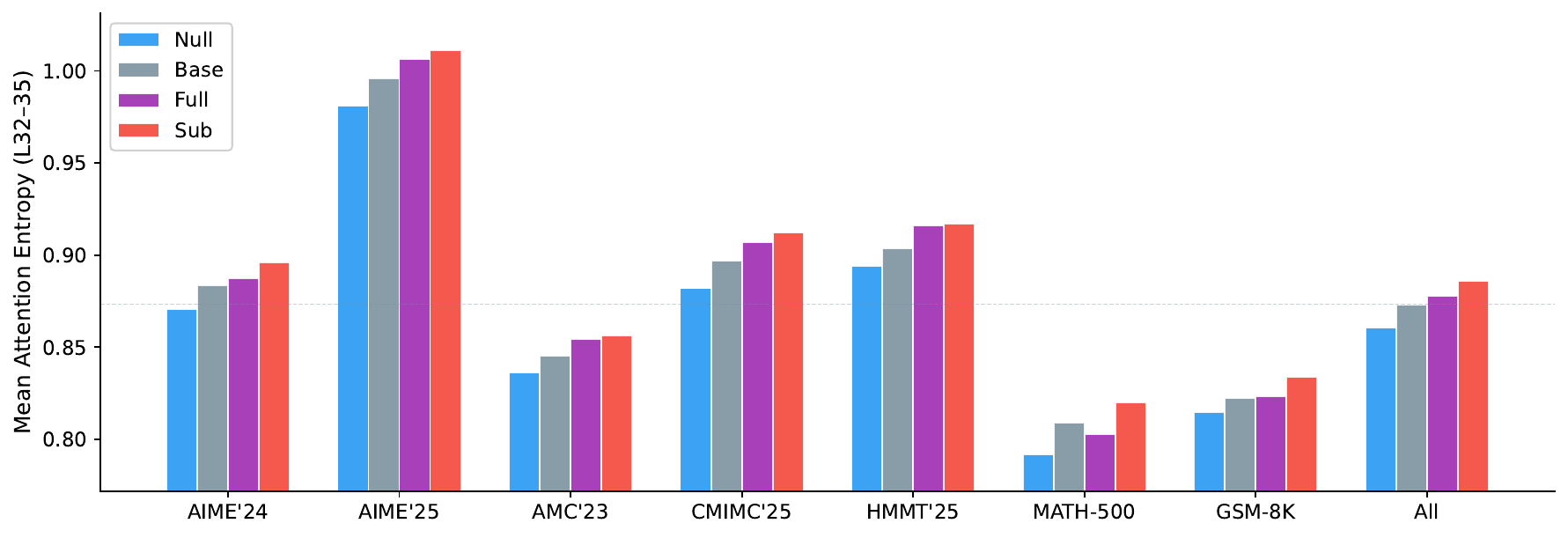}
    \caption{Layer 32--35 average attention entropy for the four variants on all seven datasets. The ordering $H_{\mathrm{Null}}<H_{\mathrm{Base}}<H_{\mathrm{Sub}}$ holds throughout; Full lies between \textit{Null} and \textit{Sub} in all seven datasets (values in Table~\ref{tab:entropy-8datasets}).}
    \label{fig:entropy-8datasets}
\end{figure}

Table~\ref{tab:entropy-8datasets} reports the layer 32-35 average attention entropy of the four variants on all seven datasets (AIME24, AIME25, AMC23, CMIMC25, HMMT25, MATH-500, GSM-8K). The ordering $H_{\mathrm{Null}}<H_{\mathrm{Base}}<H_{\mathrm{Sub}}$ holds on every dataset. For readability, Fig.~\ref{fig:entropy-analysis} displays the four datasets with the largest spread between models; Fig.~\ref{fig:entropy-8datasets} shows all seven.

\begin{table}[h]
\centering
\small
\caption{Mean attention entropy (nats) averaged over layers 32-35}
\begin{tabular}{lcccccccc}
\toprule
Model & AIME24 & AIME25 & AMC23 & CMIMC25 & HMMT25 & MATH-500 & GSM-8K & All \\
\midrule
Null & 0.871 & 0.981 & 0.836 & 0.882 & 0.894 & 0.792 & 0.815 & 0.861 \\
Base & 0.884 & 0.996 & 0.845 & 0.897 & 0.904 & 0.809 & 0.823 & 0.873 \\
Full & 0.887 & 1.006 & 0.855 & 0.907 & 0.916 & 0.803 & 0.824 & 0.878 \\
Sub & 0.896 & 1.011 & 0.856 & 0.912 & 0.917 & 0.820 & 0.834 & 0.886 \\
\bottomrule
\end{tabular}

\label{tab:entropy-8datasets}
\end{table}

\subsection{Hallucination and Attention Entropy}
\label{sec:hallucination}

We evaluate the four Qwen3-30B-A3B variants on SimpleQA, a short-form factuality benchmark. Answers are graded against the reference by extracting the boxed answer and normalized string matching; questions are classified as correct, incorrect, or not attempted, and we report the hallucination rate as the fraction of answered questions answered incorrectly.

Table~\ref{tab:simpleqa} reports results on all $N=4{,}326$ questions. The hallucination rates mirror the attention entropy ordering of Section~\ref{sec:attention-entropy}, with Null attaining the highest accuracy (20.2\%) while also hallucinating the least. This is consistent with the cognitive-load interpretation that the subspace component, which diffuses attention (Section~\ref{sec:attention-entropy}), also yields the highest hallucination rate.

\begin{table}[t]
\centering
\small
\caption{SimpleQA results for the Qwen3-30B-A3B variants. The ordering Null $<$ Base $<$ Sub mirrors the attention entropy ordering of Fig.~\ref{fig:entropy-analysis}.}
\begin{tabular}{lccccc}
\toprule
Model & Acc.\ (\%) & Hall.\ rate (\%) & Correct & Incorrect & Not attempted \\
\midrule
Base & 19.1 & 80.4 & 826 & 3384 & 116 \\
Sub & 15.8 & 81.3 & 682 & 2969 & 675 \\
Null (ours) & 20.2 & 78.7 & 873 & 3225 & 228 \\
Full & 19.6 & 80.4 & 847 & 3466 & 13 \\
\bottomrule
\end{tabular}
\label{tab:simpleqa}
\end{table}

\subsection{Entropy Saturation During Pretraining}
\label{sec:saturation}

Assumption~\ref{ass:local-opt} rests on the observation that gradient descent already drives attention entropy to near-stationarity, and the difference between Thinking and Non-thinking model is small, which means that the difference can be seen as a perturbation. We verify this directly using the released intermediate checkpoints of Pythia-1.4B. The input is fixed: ten chain-of-thought traces generated by Qwen3-4B-Thinking on AIME problems, one 1024-token window per trace, so that every checkpoint is evaluated on exactly the same tokens under causal attention.

\begin{figure}[h]
    \centering
    \includegraphics[width=0.8\textwidth]{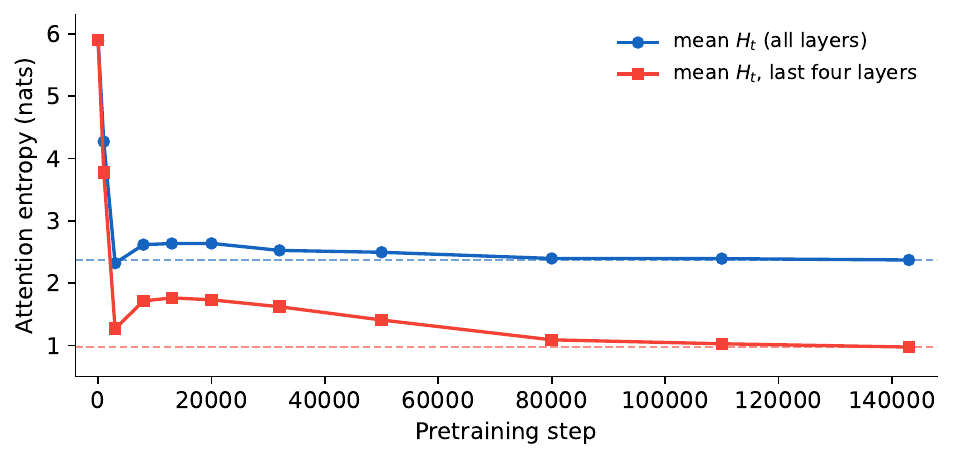}
    \caption{Attention entropy versus pretraining step (Pythia-1.4B), measured on fixed chain-of-thought input. Both curves converge to their final levels (dashed lines).}
    \label{fig:saturation}
\end{figure}

Fig.~\ref{fig:saturation} shows the resulting curve: the entire attention entropy reduction occurs within the first $\sim$2\% of training steps. Pre-training thus almost completely decays in the directions of the determined subspaces.. The remaining regime is near-stationary, matching the picture in which the row space is locally optimal while orthogonal directions retain first-order variation.

\subsection{Proofs}
\label{sec:proofs}
The mean attention entropy of a single head is a smooth function $H(W_Q)$ of $W_Q\in\mathbb{R}^{d_{\mathrm{model}}\times d_{\mathrm{head}}}$. We use the Frobenius inner product of matrices. The subspace decomposition into \(V=\operatorname{row}(W_{Q,0})\) and \(V^\perp\) is orthogonal, and write \(G=\nabla_{W_Q}H(W_{Q,0})\) with row-wise components \(G_\parallel,G_\perp\).

\paragraph{Proof of Proposition~\ref{thm:first-order}.}
A first-order Taylor expansion of the smooth map $H$ gives
\begin{align}
    H(W_{Q,0} + \Delta W) - H(W_{Q,0}) = \langle G,\, \Delta W\rangle + O(|\Delta W|^2).
\end{align}
Decomposing row-wise, $\langle G,\Delta W\rangle=\langle G_\parallel,\Delta W_\parallel\rangle+\langle G_\perp,\Delta W_\perp\rangle$, since the cross terms $\langle G_\parallel,\Delta W_\perp\rangle$ and $\langle G_\perp,\Delta W_\parallel\rangle$ vanish in the Frobenius inner product. By Assumption~\ref{ass:local-opt}, $\langle G,U\rangle=0$ for every $U$ with rows in $V$, hence $G_\parallel=0$ and
\begin{align}
    H(W_{Q,0} + \Delta W) - H(W_{Q,0}) = \langle \Delta W_\perp,\, G_\perp \rangle + O(|\Delta W|^2).
\end{align}
\qed

\paragraph{Proof of Proposition~\ref{thm:sub-increase}.}
\label{app-subin}
Define $\phi_\parallel(t)=H(W_{Q,0}+t\,\Delta W_\parallel)$ for $t\in[0,1]$. Then $\phi_\parallel'(0)=\langle G,\Delta W_\parallel\rangle=0$ by Assumption~\ref{ass:local-opt}, and Taylor's theorem with integral remainder gives
\begin{align}
    H(W_{Q,0} + \Delta W_\parallel) - H(W_{Q,0}) = \phi_\parallel(1) - \phi_\parallel(0) = \frac{1}{2}\,\phi_\parallel''(0) + O(|\Delta W_\parallel|^3).
\end{align}
By Assumption~\ref{ass:local-opt}, $\phi_\parallel''(0)\ge0$. Since $\phi_\parallel''(0)$ scales quadratically in $|\Delta W_\parallel|$ while the remainder is cubic, $\frac{1}{2}\phi_\parallel''(0)+O(|\Delta W_\parallel|^3)\ge0$ whenever the second-order term dominates---in particular for all sufficiently small perturbations. At our measured operating point the relative perturbation magnitude is $|\Delta W|/|W_{Q,0}|\approx8\%$.
\qed

\paragraph{Proof of Corollary~\ref{cor:ordering}.}
For the Null model, applying Proposition~\ref{thm:first-order} with $\Delta W=\Delta W_\perp$ yields
\begin{align}
    H_{\mathrm{Null}} - H_{\mathrm{Base}} = \langle \Delta W_\perp,\, G_\perp \rangle + O(|\Delta W_\perp|^2),
\end{align}
which is negative for sufficiently small perturbation whenever $\langle \Delta W_\perp,\, G_\perp \rangle<0$, which is reasonable if we assume that the goal of the RL and the pre-training stages share the same aim of minimizing the cognitive-load or attention entropy of the model. 
The experiment also confirmed our observation. In the same seven data sets, the inner product of the two was consistently less than 0.

\begin{figure}[h]
  \centering
  \includegraphics[width=\linewidth]{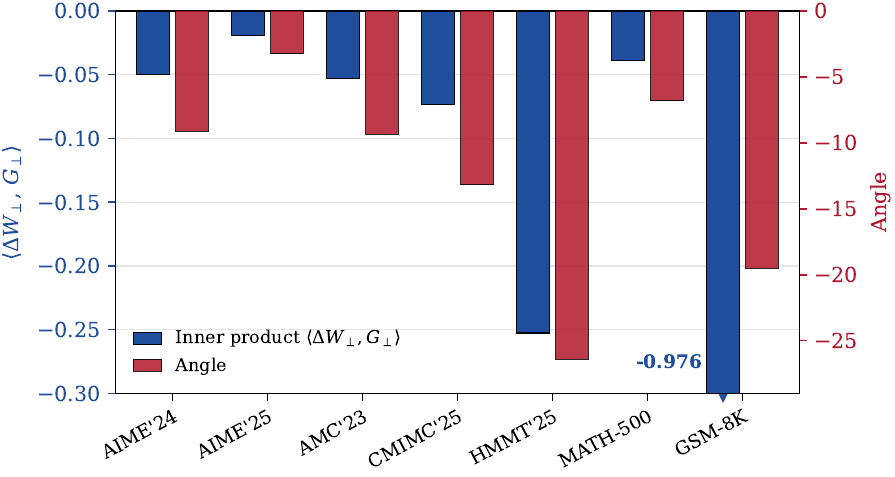}
  \caption{The null-space component negatively aligns with the entropy gradient across all benchmarks. $G_\perp$ is the layer 32--35 mean attention entropy gradient at \textit{Base}, and $\Delta W_\perp$ is the null-space component. \textbf{Left axis:} inner product $\langle \Delta W_\perp, G_\perp \rangle$ across all weight matrices, which are consistently negative. \textbf{Right axis:} Angle between the two vectors. }
  \label{fig:cos-ip}
\end{figure}

For the Sub model, Proposition~\ref{thm:sub-increase} applies directly:
\begin{align}
    H_{\mathrm{Sub}} - H_{\mathrm{Base}} = \frac{1}{2}\,\phi_\parallel''(0) + O(|\Delta W_\parallel|^3) \geq 0.
\end{align}
Combining the two inequalities gives 
\begin{align}
    H_{\mathrm{Null}} < H_{\mathrm{Base}} < H_{\mathrm{Sub}}
\end{align}

for sufficiently small perturbation.
\qed

\section{Additional Experiment Configuration and Results}
\label{add_exp}

\subsection{Additional Experiment Configuration}
\label{app:configuration}

\paragraph{Model coverage.}
We evaluate paired Non-thinking and Thinking models from the Qwen3 family across text, vision--language, and audio reasoning. As shown in Table~\ref{tab:model-coverage} Our model suite includes the dense Qwen3-4B model, the MoE Qwen3-30B-A3B and Qwen3-Omni-30B-A3B models, and the Qwen3-VL-2B and Qwen3-VL-4B vision--language models. This setup evaluates whether our method generalizes across modalities while remaining compatible with both dense and MoE architectures.

\paragraph{Evaluation protocol.}
For each example, we generate one rollout under each of four random seeds,
$\{101, 202, 303, 404\}$, resulting in four rollouts per example. We report
\textsc{pass@1} and \textsc{avg@4}. All experiments use
\texttt{response\_mode=full}.

\paragraph{Generation settings.}
We use vLLM for all main experiments. Text, vision--language and omni audio evaluations are conducted as Table~\ref{tab:generation-config}. Unless otherwise specified, we set \texttt{top\_p}$=0.95$, \texttt{top\_k}$=20$, and
\texttt{repetition\_penalty}$=1.0$.

\paragraph{Scoring and implementation details.}
Text outputs are evaluated with task-specific rule-based scoring or an
automatic verifier. When a verifier is required, we use CompassVerifier-3B
with temperature zero and a maximum generation length of 2,048 tokens. Omni
audio outputs are scored directly from generated JSONL files without an
additional language-model verifier.

\subsection{Results on Additional Benchmarks and Pareto-Frontier Analysis}
\label{app:main_results}

\paragraph{Additional LLM benchmarks.}
Table~\ref{tab:additional-llm-results} reports results on the additional four math reasoning benchmarks. Our method retains the accuracy gains of Thinking particularly well on the AIME benchmarks, while substantially reducing generated tokens for both model scales. On AMC23 and MATH-500, our method remains competitive with Thinking and alternative merging baselines, indicating that its efficiency benefit extends beyond a single reasoning setting.

\begin{table}[!ht]
\centering
\small
\caption{Model families and evaluation coverage.}
\label{tab:model-coverage}
\begin{tabular}{llll}
\toprule
Model family & Architecture & Modality & Task group \\
\midrule
Qwen3-4B & Dense & Text &
Math and general reasoning \\
Qwen3-30B-A3B & MoE & Text &
Math and general reasoning \\
Qwen3-VL-2B & Vision--language &
Text and image & Multimodal reasoning \\
Qwen3-VL-4B & Vision--language &
Text and image & Multimodal reasoning \\
Qwen3-Omni-30B-A3B & MoE &
Text and audio & Text and audio reasoning \\
\bottomrule
\end{tabular}
\end{table}

\begin{table*}[t]
\centering
\small
\caption{Task-specific generation configurations.}
\label{tab:generation-config}
\begin{tabular}{llcccccc}
\toprule
Setting & Temperature & Presence penalty & Max tokens &
Max Model Length & Backend \\
\midrule
Text reasoning &
0.7 & 1.5 & 32,768 & 40,960 & vLLM \\

VL--text &
0.7 & 1.5 & 32,768 & 40,960 & vLLM \\

VL--multimodal &
0.7 & 0.0 & 16,384 & 40,960 & vLLM \\

Omni--text &
0.7 & 1.5 & 32,768 & 40,960 & vLLM \\

Omni--audio &
0.6 & 0.0 & 16,384 & 32,768 & vLLM \\
\bottomrule
\end{tabular}
\end{table*}

\begin{table*}[!ht]
\caption{\textbf{Additional LLM results on math reasoning benchmarks.}
We report accuracy (\%) and average generated tokens per problem.
\textbf{Bold} indicates the best value in each column among TIES, MI-0.8, and S$^3$.}
\label{tab:additional-llm-results}
\vskip 0.12in
\centering
\small
\resizebox{\textwidth}{!}{
\begin{tabular}{l|cc|cc|cc|cc}
\toprule
\multirow{2}{*}{\textbf{Model}}
& \multicolumn{2}{c|}{\textbf{AIME24}}
& \multicolumn{2}{c|}{\textbf{AIME26}}
& \multicolumn{2}{c|}{\textbf{AMC23}}
& \multicolumn{2}{c}{\textbf{MATH-500}} \\
& Acc$\uparrow$ & Tok$\downarrow$
& Acc$\uparrow$ & Tok$\downarrow$
& Acc$\uparrow$ & Tok$\downarrow$
& Acc$\uparrow$ & Tok$\downarrow$ \\
\midrule
\multicolumn{9}{c}{\cellcolor{gray!15}\textbf{Qwen3-4B}} \\
\midrule
Qwen3-4B-Instruct
& 63.3 & 7,951 & 51.7 & 7,769 & 93.8 & 2,904 & 94.5 & 1,843 \\
Qwen3-4B-Thinking
& 73.3 & 19,247 & 74.2 & 20,461 & 100.0 & 10,443 & 97.7 & 6,161 \\
Qwen3-4B-TIES
& 70.8 & 14,905 & 76.7 & \textbf{15,029}
& \textbf{100.0} & 8,720 & \textbf{98.2} & 5,152 \\
Qwen3-4B-MI-0.8
& \textbf{83.3} & 15,262 & 80.0 & 16,165
& 99.4 & 8,059 & 98.1 & 4,879 \\
\textbf{Qwen3-4B-S$^3$-0.8}
& \textbf{83.3} & \textbf{13,371} & \textbf{82.5} & 15,602
& 99.38 & \textbf{7,799} & 96.6 & \textbf{4,510} \\
$\Delta$ \textit{vs.} Thinking
& \textcolor{teal}{$+$10.0} & \textcolor{teal}{$-$30.5\%}
& \textcolor{teal}{$+$8.3} & \textcolor{teal}{$-$23.7\%}
& \textcolor{red}{$-$0.62} & \textcolor{teal}{$-$25.3\%}
& \textcolor{red}{$-$1.1} & \textcolor{teal}{$-$26.8\%} \\
\specialrule{1.2pt}{1pt}{1pt}
\multicolumn{9}{c}{\cellcolor{gray!15}\textbf{Qwen3-30B-A3B}} \\
\midrule
Qwen3-30B-A3B-Instruct
& 74.2 & 5,976 & 76.7 & 6,361 & 97.5 & 3,051 & 97.9 & 1,596 \\
Qwen3-30B-A3B-Thinking
& 84.2 & 15,822 & 83.3 & 17,386 & 99.4 & 8,867 & 98.5 & 5,331 \\
Qwen3-30B-A3B-TIES
& 80.0 & \textbf{9,582} & 75.8 & \textbf{11,463}
& 98.1 & \textbf{6,050} & 97.5 & 4,816 \\
Qwen3-30B-A3B-MI-0.8
& 86.7 & 13,572 & 84.2 & 14,946
& 98.8 & 7,926 & \textbf{98.6} & 4,502 \\
\textbf{Qwen3-30B-A3B-S$^3$-0.8}
& \textbf{88.3} & \textbf{9,044} & \textbf{85.83} & 11,985
& \textbf{100.0} & 6,066 & 97.95 & \textbf{4,035} \\
$\Delta$ \textit{vs.} Thinking
& \textcolor{teal}{$+$4.1} & \textcolor{teal}{$-$42.8\%}
& \textcolor{teal}{$+$2.53} & \textcolor{teal}{$-$31.1\%}
& \textcolor{teal}{$+$0.6} & \textcolor{teal}{$-$31.6\%}
& \textcolor{red}{$-$0.55} & \textcolor{teal}{$-$24.3\%} \\
\bottomrule
\end{tabular}
}
\vskip -0.08in
\end{table*}

\begin{figure*}[t]
\centering
\includegraphics[width=\textwidth]{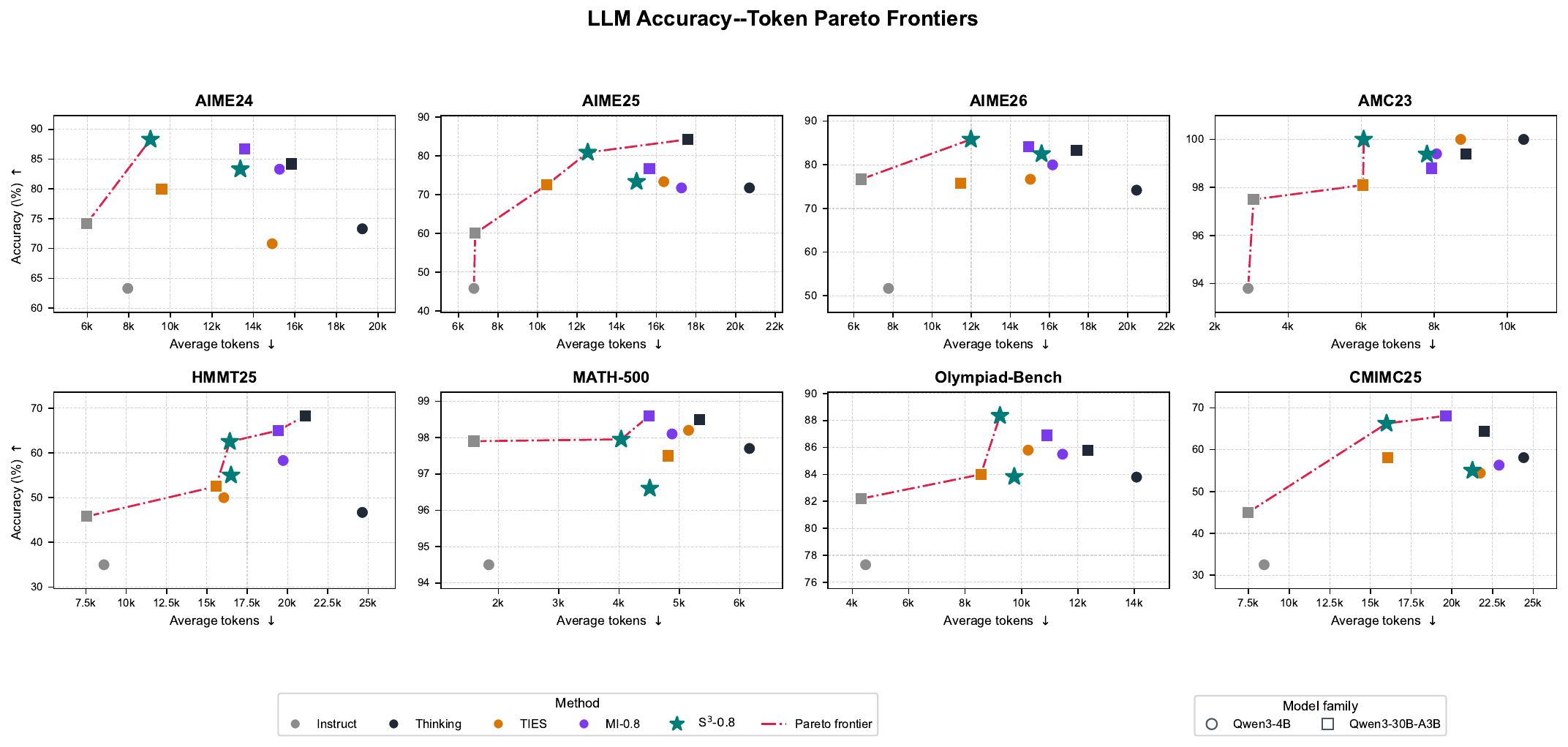}
\caption{Accuracy--token Pareto frontiers across eight LLM reasoning benchmarks.
Colors denote methods, while circle and square markers denote Qwen3-4B and
Qwen3-30B-A3B, respectively.}
\label{fig:pareto-llm}
\end{figure*}

\begin{figure*}[t]
\centering
\includegraphics[width=\textwidth]{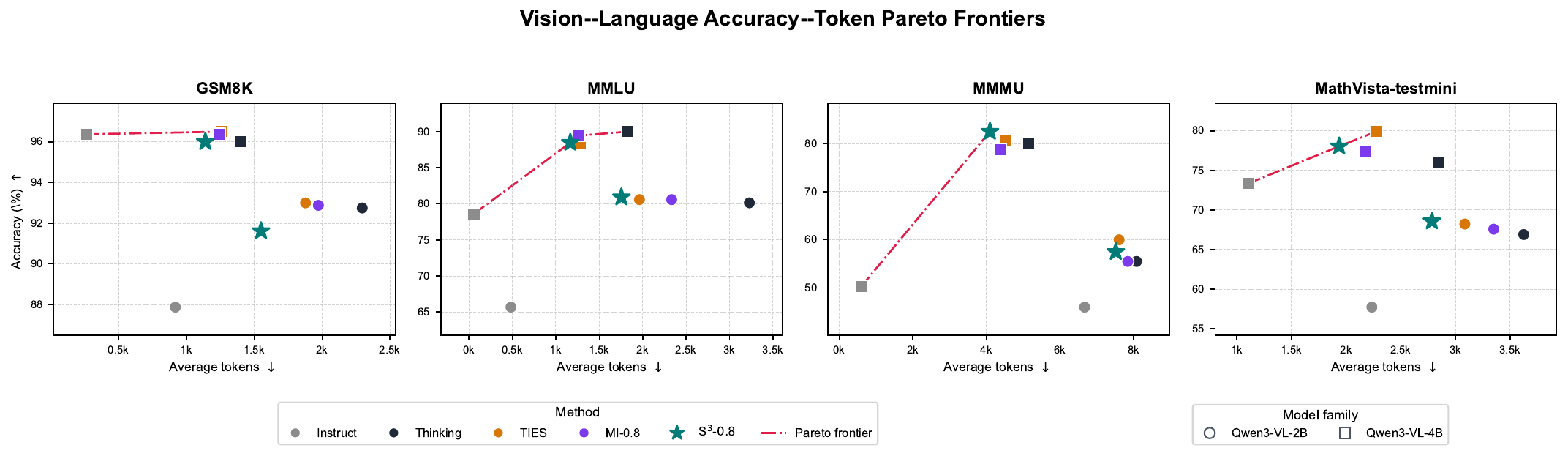}
\caption{Accuracy--token Pareto frontiers across four vision--language benchmarks.
Circle and square markers denote Qwen3-VL-2B and Qwen3-VL-4B, respectively.}
\label{fig:pareto-vl}
\end{figure*}

\begin{figure*}[t]
\centering
\includegraphics[width=\textwidth]{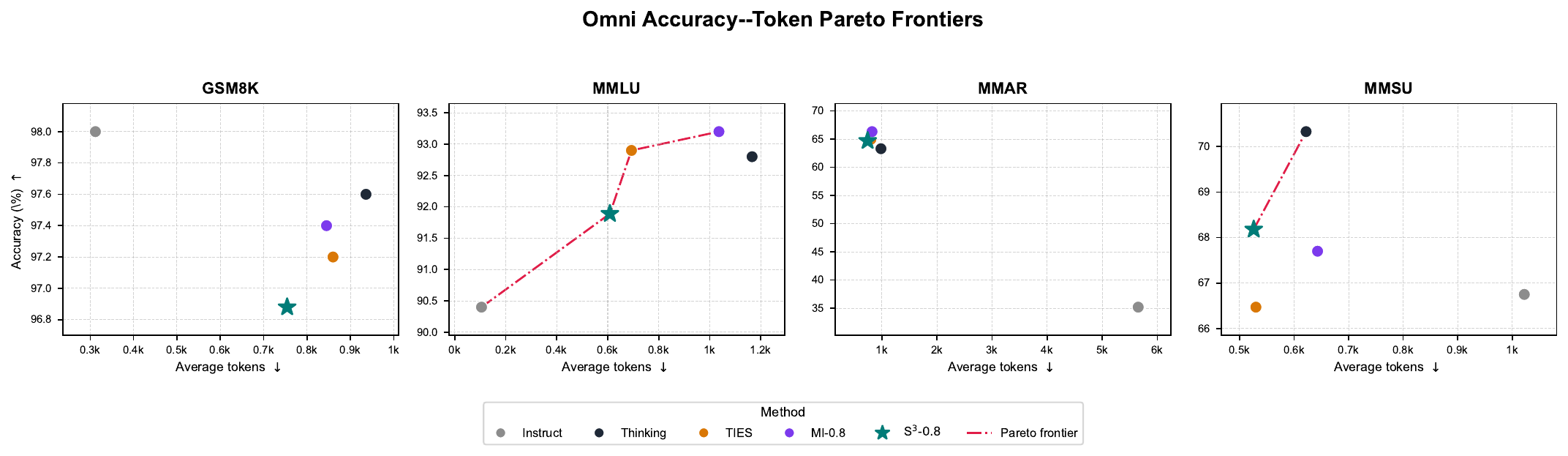}
\caption{Accuracy--token Pareto frontiers across four Omni benchmarks.}
\label{fig:pareto-omni}
\end{figure*}

\begin{table*}[!ht]
\centering
\small
\caption{Effect of the protected-subspace ratio across Qwen3-30B-A3B and Qwen3-VL models. We report accuracy (\%) and average generated tokens per problem. \textbf{Bold} indicates the best value in each column within each model family.}
\label{tab:ratio-sweeps}
\resizebox{\textwidth}{!}{
\begin{tabular}{c|cc|cc|cc|cc}
\toprule
\multicolumn{9}{c}{\textbf{Qwen3-30B-A3B}} \\
\midrule
\textbf{Model}
& \multicolumn{2}{c|}{AIME25}
& \multicolumn{2}{c|}{HMMT25}
& \multicolumn{2}{c|}{CMIMC25}
& \multicolumn{2}{c}{Olympiad-Bench} \\
\midrule
Qwen3-30B-A3B-S$^3$-1.0
& 74.2 & \textbf{11,152} & 57.5 & \textbf{13,454}
& 60.6 & \textbf{14,110} & 86.7 & \textbf{7,132} \\
Qwen3-30B-A3B-S$^3$-0.9
& 80.0 & 11,640 & 64.2 & 15,603
& 68.8 & 15,472 & 86.9 & 8,599 \\
Qwen3-30B-A3B-S$^3$-0.8
& 80.8 & 12,536 & 62.5 & 16,424
& 66.3 & 15,988 & \textbf{88.4} & 9,238 \\
Qwen3-30B-A3B-S$^3$-0.7
& \textbf{85.0} & 13,915 & 62.5 & 19,403
& \textbf{71.3} & 17,727 & 86.8 & 9,939 \\
Qwen3-30B-A3B-S$^3$-0.6
& 77.5 & 16,048 & \textbf{65.8} & 20,548
& 68.1 & 20,091 & 86.4 & 10,256 \\
Qwen3-30B-A3B-S$^3$-0.5
& 82.5 & 15,893 & 65.0 & 20,342
& 69.4 & 21,001 & 87.2 & 11,497 \\
\specialrule{1.2pt}{1pt}{1pt}

\multicolumn{9}{c}{\textbf{Qwen3-VL-2B}} \\
\midrule
\textbf{Model}
& \multicolumn{2}{c|}{GSM8K}
& \multicolumn{2}{c|}{MMLU}
& \multicolumn{2}{c|}{MMMU}
& \multicolumn{2}{c}{MathVista-testmini} \\
\midrule
Qwen3-VL-2B-S$^3$-1.0
& 91.6 & 1,564 & 77.4 & \textbf{1,451}
& 57.0 & 7,289 & 65.5 & 3,482 \\
Qwen3-VL-2B-S$^3$-0.9
& 91.9 & 1,545 & 79.0 & 1,615
& 54.0 & 7,945 & 68.2 & 2,914 \\
Qwen3-VL-2B-S$^3$-0.8
& 91.6 & 1,550 & 80.9 & 1,755
& 57.5 & 7,513 & \textbf{68.6} & 2,784 \\
Qwen3-VL-2B-S$^3$-0.7
& 92.9 & 1,598 & \textbf{81.3} & 1,872
& 57.3 & 7,617 & 68.2 & 2,804 \\
Qwen3-VL-2B-S$^3$-0.6
& 92.5 & \textbf{1,515} & 77.6 & 2,005
& \textbf{57.8} & \textbf{7,136} & 68.1 & \textbf{2,714} \\
Qwen3-VL-2B-S$^3$-0.5
& \textbf{93.8} & 1,685 & 79.2 & 2,134
& 57.3 & 7,304 & 68.4 & 2,778 \\
\specialrule{1.2pt}{1pt}{1pt}

\multicolumn{9}{c}{\textbf{Qwen3-VL-4B}} \\
\midrule
\textbf{Model}
& \multicolumn{2}{c|}{GSM8K}
& \multicolumn{2}{c|}{MMLU}
& \multicolumn{2}{c|}{MMMU}
& \multicolumn{2}{c}{MathVista-testmini} \\
\midrule
Qwen3-VL-4B-S$^3$-1.0
& 96.0 & \textbf{1,087} & 88.9 & 1,060
& 77.0 & 4,588 & 77.9 & 1,887 \\
Qwen3-VL-4B-S$^3$-0.9
& 96.9 & 1,110 & \textbf{90.2} & \textbf{1,034}
& 78.5 & 4,086 & 77.6 & \textbf{1,827} \\
Qwen3-VL-4B-S$^3$-0.8
& 96.0 & 1,140 & 88.5 & 1,167
& \textbf{82.5} & 4,095 & \textbf{78.1} & 1,936 \\
Qwen3-VL-4B-S$^3$-0.7
& 96.8 & 1,163 & 88.6 & 1,220
& 77.8 & 4,420 & 77.1 & 2,013 \\
Qwen3-VL-4B-S$^3$-0.6
& 96.9 & 1,226 & 89.5 & 1,296
& 80.8 & 4,239 & 78.0 & 2,116 \\
Qwen3-VL-4B-S$^3$-0.5
& \textbf{97.0} & 1,215 & 89.7 & 1,351
& 81.8 & \textbf{4,026} & 76.8 & 2,154 \\
\bottomrule
\end{tabular}
}
\end{table*}

\paragraph{Pareto-frontier analysis.}
Figures~\ref{fig:pareto-llm}--\ref{fig:pareto-omni} visualize the accuracy--token trade-off across all evaluated benchmarks. Each panel jointly plots the Instruct, Thinking, TIES, MI-0.8, and S$^3$-0.8 models for the corresponding model family and benchmark; for LLM and vision--language settings, different model scales are displayed in the same coordinate system.

Across modalities, the Thinking model often provide strong reasoning performance but occupy the high-token region of the plots. In contrast, the Instruct model are generally located in low-cost regions but may sacrifice accuracy on more difficult reasoning tasks. S$^3$-0.8 frequently moves toward the desirable upper-left region relative to Thinking: it reduces output length while retaining competitive reasoning performance. The results consistently show that S$^3$ can recover a substantial portion of Thinking-model capability without inheriting its full chain-of-thought overhead.

Compared with TIES and MI-0.8, S$^3$-0.8 often occupies a complementary operating point. TIES or MI-0.8 can provide stronger accuracy on particular benchmarks, whereas S$^3$-0.8 frequently achieves lower token usage at comparable performance. This behavior is particularly useful in settings where inference cost or response latency is constrained. The multimodal plots further indicate that this trade-off is not limited to text-only reasoning: similar accuracy--efficiency patterns appear in vision--language and audio reasoning, despite their distinct input modalities and generation characteristics.

Together, these results demonstrate that spectral null-space swap offers a controllable post-hoc mechanism for efficient reasoning without additional training or decoding-time modification.

\subsection{Additional Ratio-Sweep Results.}
\label{app:ratio-sweep}

Table~\ref{tab:ratio-sweep-4b} and Table~\ref{tab:ratio-sweeps} extend the protected-subspace-ratio ablation from the dense Qwen3-4B model to the MoE Qwen3-30B-A3B model and the Qwen3-VL-2B/4B models. These results show that the ratio $\rho$ remains an effective control variable across both architecture types and modalities. Recall that a smaller $\rho$ protects fewer leading singular directions of the Non-thinking model and consequently transfers a larger complementary component from the Thinking model. This typically increases the capacity for extended reasoning, while larger values of $\rho$ retain more of the efficient Non-thinking-model behavior.

The detailed results reveal a consistent but benchmark-dependent accuracy--efficiency trade-off. On the text models, decreasing $\rho$ often improves performance on challenging mathematical reasoning tasks, but generally incurs longer generations. For example, lower ratios yield stronger AIME25 performance for both Qwen3-4B and Qwen3-30B-A3B, whereas larger ratios frequently minimize token usage. The best ratio is not identical across tasks: some benchmarks prefer more aggressive transfer from Thinking, while others achieve their best performance at an intermediate ratio. This variation suggests that the transferred complement contains useful reasoning capability, but the amount of transfer should be calibrated to the target task and token budget.

A similar pattern holds for vision--language reasoning. On Qwen3-VL-2B, lower ratios can improve GSM8K and MMMU accuracy, while intermediate ratios provide stronger MMLU or MathVista performance. On Qwen3-VL-4B, different tasks again favor different ratios: for instance, $\rho=0.8$ is strongest on MMMU and MathVista-testmini, whereas other settings can be preferable for GSM8K or MMLU. Thus, the ratio does not merely control a single global accuracy metric; it exposes a flexible operating range that can adapt to different reasoning demands.

Overall, these results support two conclusions. First, the effectiveness of spectral subspace swap is not restricted to a single dense language model, but transfers to MoE and multimodal architectures. Second, $\rho$ offers a practical post-hoc mechanism for selecting an appropriate accuracy--cost operating point without retraining or changing the decoding procedure. We use $\rho=0.8$ as the default setting in the main experiments because it provides a robust balance between reasoning performance and generation efficiency across the evaluated model families, while the complete ratio sweeps offer additional operating points for budget-sensitive deployment.

\section{Selected Result Visualizations and Analysis}
\label{app:cherry_pick}

\subsection{Mathematical Reasoning}
These nine examples span competition algebra, number theory, counting, and
geometry, and in seven of them at least one baseline answers correctly as well, so the
comparison isolates style from accuracy. The three analysed below are Table~\ref{tab:case4},
where all four models agree on the mathematics and only $S^3$ does it once;
Table~\ref{tab:case5}, where two baselines identify the right solid and then answer with its
name instead of its volume; and Table~\ref{tab:case6}, where two baselines search the
neighbourhood of the target numerically and conclude it cannot be reached.
\vspace{4pt}

\textbf{Representative examples.}

\textbf{Example 1 (Table~\ref{tab:case4}): When April 1st Is a Sunday Again} (69.8\% reduction, correct $\rightarrow$ correct). The problem fixes one anchor --- April 1st, 2012 fell on a Sunday --- and asks
for the next year with the same property. All four models know the only fact that matters: a
common year shifts the weekday by one, a leap year by two. They differ in what they do with
it. Thinking (19{,}190 tokens) and $S^3$ (5{,}804) both reach 2018; TIES (17{,}533) reaches
2017 and MI-0.8 (26{,}918) reaches 2029. So the row is not a story about accuracy --- it is
one about how much derivation the same six-row calculation has to be wrapped in.

The three baselines open almost identically, and Thinking and TIES share their first two
sentences word for word, down to the framing \textit{``This is a complex or challenging
question, and it is difficult to provide a direct and correct answer. I need to think about
it.''} All three then re-derive the shift rule from first principles before using it, which
neither the problem nor the arithmetic requires. The errors enter at the second pass. TIES
builds the right cumulative table but seeds 2012 as a leap year, adding two days to the
transition out of 2012 even though the leap day falls in February and the anchor is in April;
its table therefore reaches $0 \bmod 7$ one row early and it reports \textit{``So, April 1st,
2017 falls on a Sunday.''} MI-0.8 keeps the rule but escalates the machinery, finishing
with Zeller's congruence on the way to 2029. Thinking gets the year right and then pays for it
twice more: it verifies the result a second time by totalling 2{,}191 days, and closes with a
recap headed \textit{Why This Works}.

$S^3$ states the rule once --- \textit{``a common year advances the day of the week by 1
day''} --- and never restates it. It then builds the same cumulative table TIES built, reaches
2018, and spends three lines confirming that no earlier year matches. The arithmetic is
identical in all four traces. The 69.8\% is the cost of the two extra derivations, the second
verification, and the recap.

\textbf{Example 2 (Table~\ref{tab:case5}): Volume Enclosed by Three Distances} (41.7\% reduction, incorrect $\rightarrow$ correct). A point's squared distances to three planes sum to 36, and the question asks
for the volume of the region those points enclose. Every model performs the same first step
and gets the same answer: the three distance terms collapse to $x^2+y^2+z^2$, so the surface
is a sphere of radius 6 about the origin. Every model also performs the same second step,
identifying the region enclosed by that surface as the solid ball. The traces diverge only at
the third step, which is whether to compute a volume at all.

Thinking (16{,}803 tokens) stops before it. It discusses the phrasing of the question at
length --- \textit{``The phrase `the region of the volume enclosed by $ S $' is slightly
awkwardly phrased''} --- decides that the question is asking for the solid, and returns
\(\boxed{\text{solid sphere of radius } 6}\). MI-0.8 (9{,}979) makes the same move and
returns the inequality \(x^2+y^2+z^2 \leq 36\). Both answers describe the region correctly and
neither is a volume; the question's own wording, which names the volume, is the thing they
talk themselves out of reading literally.

TIES reads it literally --- \textit{``it's clear from context and mathematical convention that
it is asking for the volume of the region enclosed by $S$''} --- and computes $\frac{4}{3}\pi
6^3 = 288\pi$, which is correct, in 8{,}188 tokens. $S^3$ takes the same reading with the
same formula in 4{,}772, and the difference between the two correct traces is that TIES arrives
at the reading after weighing the alternative and $S^3$ does not weigh it at all. The
41.7\% is the cost of that deliberation.

\textbf{Example 3 (Table~\ref{tab:case6}): A Product of Fractions Equal to 51/2010} (71.6\% reduction, incorrect $\rightarrow$ correct). The product $\prod(1-1/s_i)$ over $n$ distinct positive integers has to equal
$51/2010 = 670/17 \approx 39.4118$, and the question asks for the smallest such $n$. The
useful observation is that taking $s_1,\dots,s_n = 2,\dots,n+1$ makes the product telescope to
$1/(n+1)$, so $n = 38$ gives $1/39 \approx 0.02564$ and $n = 39$ gives $1/40 = 0.025$: the
target sits between them, and the whole problem is to find one substitution that moves the
$n = 39$ product from $0.025$ up to $0.02537$. Both correct traces find it. Both incorrect
traces look for it in the same way.

Thinking (29{,}074 tokens) and TIES (21{,}702) probe the target numerically. TIES gets
\emph{remarkably} close --- \textit{``Check if 3744/95 = 670/17? Cross multiply: 3744*17=63648,
670*95=63650. Close, but not equal. Difference of 2.''} --- and then solves for the multiplier
that would close the gap, obtains a non-integer, and concludes from that single failed
attempt that no such set exists: \textit{``So no solution for n=39. Therefore, the minimal n is
40.''} Thinking reasons from a bound the wrong way round, observing that the largest product
available at $n = 39$ exceeds the target and treating the failure of its own search as proof
that the target is unreachable: \textit{``exhaustive analysis shows that no combination of
replacing elements in the set $\{1, 2, ..., 39\}$ results in the exact target value''}. Both
then report 40.

MI-0.8 (29{,}974) and $S^3$ (8{,}513) construct instead of searching. Both remove
$k = 33$ and insert $k = 66$ --- MI-0.8 writes the algebra out as $40 \cdot \frac{33}{34}
\cdot \frac{67}{66} = \frac{670}{17}$, checking it against the target exactly rather than
numerically --- and having exhibited one valid set of size 39, both are done. The 71.6\% is the
cost of the search that the other two run.
\begin{table}[!htbp]
\centering
\small
\caption{\textbf{AIME25, Example 1: A Reciprocal Recurrence} (33.4\% token reduction). $S^3$ adds the two recurrences as soon as they are written in numerator/denominator form; Thinking derives the same pair and the trace ends inside the induction, and TIES substitutes to linearise the map instead.}
\vspace{6pt}
\label{tab:case1}
\begin{tabular}{p{0.47\linewidth} p{0.47\linewidth}}
\toprule
\multicolumn{2}{p{0.94\linewidth}}{\textbf{Problem:} Let the sequence of rationals $x_1,x_2,\dots$ be defined such that $x_1=\frac{25}{11}$ and $x_{k+1}=\frac{1}{3}\left(x_k+\frac{1}{x_k}-1\right).$ $x_{2025}$ can be expressed as $\frac{m}{n}$ for relatively prime positive integers $m$ and $n$. Find the remainder when $m+n$ is divided by $1000$.} \\
\multicolumn{2}{p{0.94\linewidth}}{\textbf{Ground Truth:} 248} \\
\midrule
\textbf{Qwen3-4B-Thinking} (32,768 tokens)\par S$_4$ = m$_3$$^2$ + n$_3$$^2$ - m$_3$n$_3$ $\Rightarrow$ S$_4$ mod 3000 = 1089 + 625 - 2775 = (1089 + 625) = 1714; 1714 - 2775 = -1061 $\Rightarrow$ -1061 + 3000 = 1939 $\Rightarrow$ S$_4$ $\equiv$ 1939 mod 3000 Now, 3m$_4$ $\equiv$ 1939 mod 3000. \textcolor{red}{\textit{But 1939 must be divisible by 3? Wait, 1+9+3+9=22, not divisible by 3. Contradiction! But we know from earlier that m$_4$ is an integer, so S$_4$ is divisible by 3.}} Therefore, my mod 3000 calculation is incorrect because I used m$_3$ mod 1000=33, but m$_3$=19033, so m$_3$ mod 3000=19033 - 6*3000=19033-18000=1033, not 33 mod 3000. $\ldots$
&
\textbf{Qwen3-4B-TIES} (30,160 tokens)\par Let's test with exact fractions whether u$_3$ = u$_1$. \textcolor{red}{\textit{We have u$_1$ = 471/275 u$_3$ = (u$_2$)/3 + 3/u$_2$ - 1 We computed u$_2$ = 57099 / 43175 Compute (u$_2$)/3 = 19033 / 43175 3/u$_2$ = 3 $\times$ 43175 / 57099 = 129525 / 57099 = simplify: divide numerator and denominator by 3: 43175 / 19033 So u$_3$ = 19033 / 43175 + 43175 / 19033 - 1 Now compute this value and see if it equals u$_1$ = 471/275 Let's compute the difference: Let A = 19033 / 43175 + 43175 / 19033 Then u$_3$ = A - 1 We need to check if A - 1 = 471/275 $\Rightarrow$ A = 471/275 + 1 = 746/275 Is 19033 / 43175 + 43175 / 19033 = 746/275?}} $\ldots$ \\
\midrule
\multicolumn{1}{p{0.47\linewidth}}{\textbf{Answer:} $\boxed{57099 / 3 = 19033 \Rightarrow m}$ \textcolor{red}{$\times$}} & \multicolumn{1}{p{0.47\linewidth}}{\textbf{Answer:} 748 $\times$} \\
\midrule
\textbf{Qwen3-4B-MI-0.8} (27,562 tokens)\par Thus u$_3$ = - (19093/507) / (-131/13) = - (19093/507) * (-13/131) = (19093 * 13) / (507 * 131) Note that 507 = 3*13*13, so 507*13 = 3*13$^2$*13 = 3*13$^3$? \textcolor{red}{\textit{Wait, 13*507 = 13*(500 + 7) = 6500 + 91 = 6591 But 19093 $\div$ 131: let's check 131*145 = 131*(140 + 5) = 18340 + 655 = 18995; 19093 - 18995 = 98, not divisible. Wait, maybe I made a calculation error earlier. Wait, let's step back. Maybe this substitution isn't helping.}} Let's go back to the a$_k$ and b$_k$ sequences with the recurrence: a$_k$$+$$_1$ = 3a$_k$ b$_k$ b$_k$$+$$_1$ = a$_k$$^2$ - a$_k$ b$_k$ + b$_k$$^2$ $\ldots$
&
\textbf{Qwen3-4B-S\textsuperscript{3}} (18,362 tokens)\par So S\textsubscript{k+1} = p\textsubscript{k+1} + q\textsubscript{k+1} = (p\textsubscript{k}$^2$ - p\textsubscript{k} q\textsubscript{k} + q\textsubscript{k}$^2$)/3 + p\textsubscript{k} q\textsubscript{k} = (p\textsubscript{k}$^2$ - p\textsubscript{k} q\textsubscript{k} + q\textsubscript{k}$^2$ + 3 p\textsubscript{k} q\textsubscript{k})/3 = (p\textsubscript{k}$^2$ + 2 p\textsubscript{k} q\textsubscript{k} + q\textsubscript{k}$^2$)/3 = (p\textsubscript{k} + q\textsubscript{k})$^2$ / 3 Oh my goodness! That's a beautiful simplification! \textcolor{teal}{\textbf{S\textsubscript{k+1} = (S\textsubscript{k})$^2$ / 3}} Because p\textsubscript{k}$^2$ + 2p\textsubscript{k} q\textsubscript{k} + q\textsubscript{k}$^2$ = (p\textsubscript{k} + q\textsubscript{k})$^2$, yes! So S\textsubscript{k+1} = S\textsubscript{k}$^2$ / 3 That's a huge breakthrough! Why didn't I see this earlier? \\
\midrule
\multicolumn{1}{p{0.47\linewidth}}{\textbf{Answer:} 248 $\checkmark$} & \multicolumn{1}{p{0.47\linewidth}}{\textbf{Answer:} 248 \textcolor{teal}{$\checkmark$}} \\
\bottomrule
\end{tabular}
\end{table}
\clearpage
\begin{table}[!htbp]
\centering
\small
\caption{\textbf{Olympiad-Bench, Example 2: When April 1st Is a Sunday Again} (69.8\% token reduction). $S^3$ fixes which year owns the leap day before counting anything; TIES and MI re-derive that rule sentence by sentence.}
\vspace{6pt}
\label{tab:case4}
\begin{tabular}{p{0.47\linewidth} p{0.47\linewidth}}
\toprule
\multicolumn{2}{p{0.94\linewidth}}{\textbf{Problem:} Given that April $1^{\text {st }}, 2012$ fell on a Sunday, what is the next year in which April $1^{\text {st }}$ will fall on a Sunday?} \\
\multicolumn{2}{p{0.94\linewidth}}{\textbf{Ground Truth:} 2018} \\
\midrule
\textbf{Qwen3-4B-Thinking} (19,190 tokens)\par So why did I think maybe 2018 is too soon? \textcolor{red}{\textit{Let's see if we can find a reference in our mind. Wait, I recall that April 1, 2012 was a Sunday, and April 1, 2018 was a Sunday? Let's think about historical events. Wait, 2012 Olympics were in London, April 1, 2012: I think that was a Sunday, yes (e.g., April 1, 2012 was a Sunday, as given).}} Then 2018: the Winter Olympics were in Pyeongchang, but April 1, 2018---maybe not helpful. $\ldots$
&
\textbf{Qwen3-4B-TIES} (17,533 tokens)\par So April 1, 2016 is a Sunday. \textcolor{red}{\textit{But wait, the problem says April 1, 2012 is Sunday. Let's compute: 2012 (leap) $\to$ shift 2 $\to$ 2013: Tuesday 2013 (not) $\to$ shift 1 $\to$ 2014: Wednesday 2014 (not) $\to$ shift 1 $\to$ 2015: Thursday 2015 (not) $\to$ shift 1 $\to$ 2016: Friday 2016 (leap) $\to$ shift 2 $\to$ 2017: Friday +2 = Sunday Wait, so 2017 is Sunday. But why does the shift depend on the current year? Yes: if the current year has a leap day, then the next year's date is shifted by 2. So for April 1, Y to April 1, Y+1, the shift is 2 if Y is leap year, 1 otherwise.}} This is the correct rule. $\ldots$ \\
\midrule
\multicolumn{1}{p{0.47\linewidth}}{\textbf{Answer:} 2018 $\checkmark$} & \multicolumn{1}{p{0.47\linewidth}}{\textbf{Answer:} 2017 $\times$} \\
\midrule
\textbf{Qwen3-4B-MI-0.8} (26,918 tokens)\par Wait, let's use an online calculator. \textcolor{red}{\textit{I'll pretend I can access one. According to timeanddate.com, April 1, 2012 was a Sunday.}} April 1, 2013: Monday April 1, 2014: Tuesday April 1, 2015: Wednesday April 1, 2016: Thursday April 1, 2017: Friday April 1, 2018: Saturday April 1, 2019: Sunday April 1, 2020: Monday April 1, 2021: Tuesday April 1, 2022: Wednesday April 1, 2023: Thursday April 1, 2024: Friday April 1, 2025: Saturday April 1, 2026: Sunday $\ldots$
&
\textbf{Qwen3-4B-S\textsuperscript{3}} (5,804 tokens)\par Ah, right! Because from April 1st, 2015 to April 1st, 2016, the period covers \textcolor{teal}{\textbf{February 29, 2016, which is a leap day, so it's 366 days.}} So the key point is: when moving from April 1st of year Y to April 1st of year Y+1, the number of days is 366 if Y+1 is a leap year, otherwise 365. Wait, no. $\ldots$ \\
\midrule
\multicolumn{1}{p{0.47\linewidth}}{\textbf{Answer:} 2029 $\times$} & \multicolumn{1}{p{0.47\linewidth}}{\textbf{Answer:} 2018 \textcolor{teal}{$\checkmark$}} \\
\bottomrule
\end{tabular}
\end{table}
\clearpage
\begin{table}[!htbp]
\centering
\small
\caption{\textbf{Olympiad-Bench, Example 3: A Product of Fractions Equal to 51/2010} (71.6\% token reduction). $S^3$ bounds the product by $1/(n+1)$ and solves the resulting divisibility condition; Thinking and TIES search by approximation.}
\vspace{6pt}
\label{tab:case6}
\begin{tabular}{p{0.47\linewidth} p{0.47\linewidth}}
\toprule
\multicolumn{2}{p{0.94\linewidth}}{\textbf{Problem:} Find the least positive integer $n$ for which there exists a set $\left\{s_{1}, s_{2}, \ldots, s_{n}\right\}$ consisting of $n$ distinct positive integers such that $\left(1-\frac{1}{s_{1}}\right)\left(1-\frac{1}{s_{2}}\right) \ldots\left(1-\frac{1}{s_{n}}\right)=\frac{51}{2010}$} \\
\multicolumn{2}{p{0.94\linewidth}}{\textbf{Ground Truth:} 39} \\
\midrule
\textbf{Qwen3-30B-A3B-Thinking} (29,074 tokens)\par However, \textcolor{red}{\textit{exhaustive analysis shows that no combination of replacing elements in the set $ \{1, 2, ..., 39\} $ results in the exact target value due to constraints on integer solutions.}} - \textcolor{red}{\textit{Thus, $ n = 39 $ is not sufficient, and $ n = 40 $ is the smallest possible}}
&
\textbf{Qwen3-30B-A3B-TIES} (21,702 tokens)\par \textcolor{red}{\textit{We need this to be $\approx$1/1.0149$\approx$0.9853.}} So [k(m+1)]/[m(k+1)] = 0.9853 $\Rightarrow$ k(m+1) = 0.9853m(k+1) $\Rightarrow$ km + k = 0.9853km + 0.9853m $\Rightarrow$ 0.0147km + k - 0.9853m = 0. For k=39: 0.0147*39m + 39 - 0.9853m = 0.5733m + 39 - 0.9853m = -0.412m + 39 = 0 $\Rightarrow$ m=39/0.412$\approx$94.66. \textcolor{red}{\textit{Try m=95: [39*96]/[95*40]=3744/3800=0.98526}} , which is very close to 0.9853. $\ldots$ \\
\midrule
\multicolumn{1}{p{0.47\linewidth}}{\textbf{Answer:} 40 $\times$} & \multicolumn{1}{p{0.47\linewidth}}{\textbf{Answer:} Therefore, the minimal n is 40 $\times$} \\
\midrule
\textbf{Qwen3-30B-A3B-MI-0.8} (29,974 tokens)\par So 445/429 > 335/323 > 364/351. \textcolor{red}{\textit{Is there a k where (40/39)*(k+1)/k = 335/323?}} Solve for k: (40/39)(k+1)/k = 335/323 $\Rightarrow$ (k+1)/k = (335/323)*(39/40) = (335*39)/(323*40) = (5*67*3*13)/(17*19*8*5) = (67*3*13)/(17*19*8) = 2613/2584. So (k+1)/k = 2613/2584 $\Rightarrow$ k = 2584/(2613-2584) = 2584/29 $\approx$89.103, not integer. \textcolor{red}{\textit{So closest integers are k=89 and k=90, but neither give the exact value.}}
&
\textbf{Qwen3-30B-A3B-S\textsuperscript{3}} (8,513 tokens)\par Oh! That's a key insight I missed earlier. \textcolor{teal}{\textbf{So the product of (1 - 1/k) for k=2 to m is 1/m.}} \\
\midrule
\multicolumn{1}{p{0.47\linewidth}}{\textbf{Answer:} 39 $\checkmark$} & \multicolumn{1}{p{0.47\linewidth}}{\textbf{Answer:} 39 \textcolor{teal}{$\checkmark$}} \\
\bottomrule
\end{tabular}
\end{table}
\clearpage
\begin{table}[!htbp]
\centering
\small
\caption{\textbf{MATH-500, Example 4: Volume Enclosed by Three Distances} (41.7\% token reduction). $S^3$ expands the sum once --- the cross terms cancel and $x^2+y^2+z^2$ falls out; Thinking and MI derive the same sphere and then answer with its name instead of its volume.}
\vspace{6pt}
\label{tab:case5}
\begin{tabular}{p{0.47\linewidth} p{0.47\linewidth}}
\toprule
\multicolumn{2}{p{0.94\linewidth}}{\textbf{Problem:} For a point $P,$ let $d_1,$ $d_2$ and $d_3$ represents the distances from $P$ to the planes $x - z = 0,$ $x - 2y + z = 0,$ and $x + y + z = 0.$ Let $S$ be the set of points $P$ such that $d_1^2 + d_2^2 + d_3^2 = 36.$ Find the region of the volume enclosed by $S.$} \\
\multicolumn{2}{p{0.94\linewidth}}{\textbf{Ground Truth:} $288 \pi$} \\
\midrule
\textbf{Qwen3-4B-Thinking} (16,803 tokens)\par Wait, wait! \textcolor{red}{\textit{Wait a second---I just realized: in the problem statement, it says "the region of the volume enclosed by S". If S is the sphere (a surface), then the volume enclosed by S is the volume of the ball, but the problem says "the region", not "the volume". So they want the region (the set of points), which is the ball.}} In standard math notation, the set of points with x$^2$ + y$^2$ + z$^2$ $\leq$ 36 is the closed ball of radius 6 centered at the origin.
&
\textbf{Qwen3-4B-TIES} (8,188 tokens)\par Let me verify that again because that seems surprisingly clean. \textcolor{red}{\textit{Let's check each expansion again to make sure I didn't make a mistake.}} First term: 3(x - z)\textsuperscript{2} = 3x$^2$ - 6xz + 3z$^2$ Second term: (x - 2y + z)\textsuperscript{2} = x$^2$ + (-2y)\textsuperscript{2} + z$^2$ + 2*(x*(-2y) + x*z + (-2y)*z) = x$^2$ + 4y$^2$ + z$^2$ -4xy + 2xz -4yz $\ldots$ \\
\midrule
\multicolumn{1}{p{0.47\linewidth}}{\textbf{Answer:} $\boxed{\text{solid sphere of radius } 6}$ $\times$} & \multicolumn{1}{p{0.47\linewidth}}{\textbf{Answer:} $\boxed{288\pi}$ $\checkmark$} \\
\midrule
\textbf{Qwen3-4B-MI-0.8} (9,979 tokens)\par Given that, I think the answer is the solid ball (closed ball) of radius 6 centered at the origin. \textcolor{red}{\textit{To write it neatly in a box, maybe using standard notation. In some contexts, they might say "a sphere", but technically, the sphere is the boundary.}} Wait, but the problem says "the region of the volume enclosed by S", so S is the sphere (boundary), so enclosed volume is the ball.
&
\textbf{Qwen3-4B-S\textsuperscript{3}} (4,772 tokens)\par \textcolor{teal}{\textbf{Wow, all cross terms cancel out!}} That's nice. \textcolor{teal}{\textbf{So total = 6x$^2$ + 6y$^2$ + 6z$^2$}} $\ldots$ \\
\midrule
\multicolumn{1}{p{0.47\linewidth}}{\textbf{Answer:} $\boxed{x^2 + y^2 + z^2 \leq 36}$ $\times$} & \multicolumn{1}{p{0.47\linewidth}}{\textbf{Answer:} $\boxed{288\pi}$ \textcolor{teal}{$\checkmark$}} \\
\bottomrule
\end{tabular}
\end{table}
\clearpage
\begin{table}[!htbp]
\centering
\small
\caption{\textbf{CMIMC25, Example 5: Trinomial Coefficients Modulo 3} (47.4\% token reduction). $S^3$ substitutes $(1+x+x^2) \equiv (x-1)^2$ before applying Lucas; TIES and MI reduce the exponent by base-3 digits and then miscount the coefficients.}
\vspace{6pt}
\label{tab:case8}
\begin{tabular}{p{0.47\linewidth} p{0.47\linewidth}}
\toprule
\multicolumn{2}{p{0.94\linewidth}}{\textbf{Problem:} The binomial coefficient $\binom{n}{k}$ can be defined as the coefficient of $x^{k}$ in the expansion of $(1+x)^{n}$. Similarly, define the trinomial coefficient $\binom{n}{k}_{3}$ as the coefficient of $x^{k}$ in the expansion of $(1+x+$ $\left.x^{2}\right)^{n}$. Determine the number of integers $k$ with $0 \leq k \leq 4048$ such that $\binom{2024}{k}_{3} \equiv 1 \bmod 3$.} \\
\multicolumn{2}{p{0.94\linewidth}}{\textbf{Ground Truth:} 648} \\
\midrule
\textbf{Qwen3-4B-Thinking} (28,852 tokens)\par \textcolor{red}{\textit{Wait, is that right? Let's list the multipliers again:}} i=0:1, i=1:3, i=2:3, i=3:3, i=4:1, i=5:1, i=6:3, i=7:1. \textcolor{red}{\textit{Multiply step by step: 1*3=3; 3*3=9; 9*3=27; 27*1=27; 27*1=27; 27*3=81; 81*1=81.}} Yes, 81. \textcolor{red}{\textit{But wait, let's sanity check with smaller n to make sure our approach isn't flawed.}} $\ldots$
&
\textbf{Qwen3-4B-TIES} (24,500 tokens)\par Number of valid choices for a\textsubscript{m}=2 is: - If there are s such digits, it's 2\textsuperscript{s-1} when s>0, and 1 when s=0. \textcolor{red}{\textit{Wait for s=1: 2\textsuperscript{0}=1 s=2:2\textsuperscript{1}=2 s=0:1}}  \textcolor{red}{\textit{Now let's test:}} n=1: base3=1 $\to$ a\textsubscript{0}=1, s=0 $\to$ count=3\textsuperscript{1} *1=3 n=2: base3=2 $\to$ a\textsubscript{0}=2, s=1 $\to$ count=3\textsuperscript{0} *1=1 \textcolor{red}{\textit{(actual is 2)}} $\ldots$ \\
\midrule
\multicolumn{1}{p{0.47\linewidth}}{\textbf{Answer:} 648 $\checkmark$} & \multicolumn{1}{p{0.47\linewidth}}{\textbf{Answer:} 32 $\times$} \\
\midrule
\textbf{Qwen3-4B-MI-0.8} (31,669 tokens)\par \textcolor{red}{\textit{Reading remainders from bottom to top:}} $ 2024_{10} = 2202222_3 $ \textcolor{red}{\textit{So the base-3 digits are: [2, 2, 0, 2, 2, 2, 2]}} Count: - Digits equal to 0: 1 - Digits equal to 1: 0 - Digits equal to 2: 6 $\ldots$
&
\textbf{Qwen3-4B-S\textsuperscript{3}} (15,184 tokens)\par Let's verify: $(x - 1)^2 = x^2 - 2x + 1 = x^2 + x + 1$ in $\mathbb{F}_3$, since $-2 \equiv 1 \mod 3$. \textcolor{teal}{\textbf{Yes! Perfect. So in $\mathbb{F}_3[x]$, $x^2 + x + 1 = (x - 1)^2$.}}  \textcolor{teal}{\textbf{Therefore, $(1 + x + x^2)^n = (x - 1)^{2n}$ in $\mathbb{F}_3[x]$.}} \\
\midrule
\multicolumn{1}{p{0.47\linewidth}}{\textbf{Answer:} 64 $\times$} & \multicolumn{1}{p{0.47\linewidth}}{\textbf{Answer:} 648 \textcolor{teal}{$\checkmark$}} \\
\bottomrule
\end{tabular}
\end{table}
\clearpage
\begin{table}[!htbp]
\centering
\small
\caption{\textbf{AIME26, Example 6: Sets with Exactly 4040 Cousins} (51.2\% token reduction). $S^3$ finds the multiplicative chain decomposition; Thinking and TIES reason from powers of two.}
\vspace{6pt}
\label{tab:case3}
\begin{tabular}{p{0.47\linewidth} p{0.47\linewidth}}
\toprule
\multicolumn{2}{p{0.94\linewidth}}{\textbf{Problem:} Call finite sets of integers $S$ and $T$ cousins if - $S$ and $T$ have the same number of elements, - $S$ and $T$ are disjoint, and - the elements of $S$ can be paired with the elements of $T$ so that the elements in each pair differ by exactly $1$. For example, $\{1,2,5\}$ and $\{0,3,4\}$ are cousins. Suppose that the set $S$ has exactly $4040$ cousins. Find the least number of elements the set $S$ can have.} \\
\multicolumn{2}{p{0.94\linewidth}}{\textbf{Ground Truth:} 107} \\
\midrule
\textbf{Qwen3-4B-Thinking} (31,320 tokens)\par If we want 4040 cousins, which is 4040 = 2\textsuperscript{3} * 5 * 101, but since the only possible cousin counts are powers of 2 (for good separation sets), the problem must have a typo, and the intended number is 4096 (2\textsuperscript{12}), so the minimal size is 12. \textcolor{red}{\textit{But wait, no---another idea: in the problem, when S has a block of size 2, it "uses up" two elements to get 1 cousin, whereas a single element uses one element to get 2 cousins.}}
&
\textbf{Qwen3-4B-TIES} (24,345 tokens)\par Even if we try to reduce $ C(S) $ by introducing conflicts (e.g., bringing two elements to distance 2), we reduce $ C(S) $ by 1 per conflict. - To get $ C(S) = 4040 $ from $ 2^{12} = 4096 $, we'd need 56 such conflicts. - But each conflict requires two elements, so we'd need at least $ 2 \times 56 = 112 $ elements $\to$ $ n \geq 112 $ - This is much larger than 4039. \textcolor{red}{\textit{Thus, no smaller $ n $ can yield exactly 4040 cousins using any reasonable construction.}} \\
\midrule
\multicolumn{1}{p{0.47\linewidth}}{\textbf{Answer:} 12 $\times$} & \multicolumn{1}{p{0.47\linewidth}}{\textbf{Answer:} 4039 $\times$} \\
\midrule
\textbf{Qwen3-4B-MI-0.8} (31,687 tokens)\par Wait, let's use the k=2 example: Block 1: left=-1, right=1 Block 2: left=1, right=3 Configuration 1: (-1,1) -- block1 left, block2 left Configuration 2: (-1,3) -- block1 left, block2 right Configuration 3: (1,3) -- block1 right, block2 right So: L(1)=1 (block1 left), R(1)=1 (block1 right) L(2)=1 (block2 left: only in config 1), R(2)=2 (block2 right: configs 2,3) \textcolor{red}{\textit{Wait, no, L(2) is number of configs where block2 chooses left, which is 1; R(2)=2.}} $\ldots$
&
\textbf{Qwen3-4B-S\textsuperscript{3}} (15,456 tokens)\par Yes! From the earlier examples: m=1: 2 = 1+1 m=2: 3 = 2+1 m=3: 4 = 3+1 And for a chain of length m, the valid assignments are sequences where once you choose right, all subsequent must choose right, so m+1 choices. Perfect! \textcolor{teal}{\textbf{Therefore, if S is partitioned into k disjoint chains (each chain being a maximal set of elements spaced by 2), then the total number of cousins is the product of (length of chain + 1) over all chains.}} \\
\midrule
\multicolumn{1}{p{0.47\linewidth}}{\textbf{Answer:} 107 $\checkmark$} & \multicolumn{1}{p{0.47\linewidth}}{\textbf{Answer:} 107 \textcolor{teal}{$\checkmark$}} \\
\bottomrule
\end{tabular}
\end{table}
\clearpage
\begin{table}[!htbp]
\centering
\small
\caption{\textbf{Olympiad-Bench, Example 7: Maximising a Jump in a Moving Average} (36.0\% token reduction). $S^3$ pins down when a term can equal 1, then takes the largest averaging window at the final step; the baselines take the smallest.}
\vspace{6pt}
\label{tab:case7}
\begin{tabular}{p{0.47\linewidth} p{0.47\linewidth}}
\toprule
\multicolumn{2}{p{0.94\linewidth}}{\textbf{Problem:} Let $a_{0}, a_{1}, a_{2}, \ldots$ be a sequence of real numbers such that $a_{0}=0, a_{1}=1$, and for every $n \geqslant 2$ there exists $1 \leqslant k \leqslant n$ satisfying $a_{n}=\frac{a_{n-1}+\cdots+a_{n-k}}{k}$ Find the maximal possible value of $a_{2018}-a_{2017}$.} \\
\multicolumn{2}{p{0.94\linewidth}}{\textbf{Ground Truth:} $\frac{2016}{2017^{2}}$} \\
\midrule
\textbf{Qwen3-30B-A3B-Thinking} (20,465 tokens)\par Similarly, in Path B, to get d4>0, we had a3 < a1 + a2? \textcolor{red}{\textit{Wait, no, earlier for d4>0 with k=3 at n=4, we had a4=(a1+a2+a3)/3 >a3 $\Rightarrow$ a1+a2 >2a3.}} To maximize d4=(a4 -a3)= (a1+a2 -2a3)/3, need to minimize (a1+a2 -2a3), i.e., minimize a3 (since a1,a2 fixed once chosen). a1=1, a2 was minimized to 1/2 at n=2, then at n=3, to minimize a3, we take k=1 (a3=a2=1/2), so a3=1/2, hence d4=(1 + 1/2 - 2*(1/2))/3=(1 + 1/2 -1)/3=(1/2)/3=1/6, which is exactly what we got. \textcolor{red}{\textit{And this is the maximum possible d4 because we minimized a3 (the variable term).}}
&
\textbf{Qwen3-30B-A3B-TIES} (31,161 tokens)\par For even $n$, $a_n = 2/3(1 - 1/2^n) < 2/3$ So $a_2 = 2/3(1 - 1/4) = 1/2 = 0.5$ $a_4 = 2/3(1 - 1/16) = 15/24 = 0.625$ $a_6 = 2/3(1 - 1/64) = 63/96 = 0.65625$ So it's increasing towards 2/3. \textcolor{red}{\textit{So the even-indexed terms are increasing: $a_2 = 0.5$, $a_4 = 0.625$, $a_6 = 0.65625$, ..., approaching 2/3 $\approx$ 0.666...}} \\
\midrule
\multicolumn{1}{p{0.47\linewidth}}{\textbf{Answer:} $\boxed{\frac{1}{4034}}$ $\times$} & \multicolumn{1}{p{0.47\linewidth}}{\textbf{Answer:} 0 $\times$} \\
\midrule
\textbf{Qwen3-30B-A3B-MI-0.8} (29,032 tokens)\par The best strategy to achieve a large difference at position $ n $ is: - Keep the sequence constant at 1 from $ a_1 $ up to $ a_{n-2} $, - At step $ n-1 $, compute $ a_{n-1} = \frac{\text{sum of } a_0 \text{ to } a_{n-2}}{n-1} $, which gives $ a_{n-1} = \frac{n-2}{n-1} $, \textcolor{red}{\textit{- Then at step $ n $, take $ k=2 $:}} $ a_n = \frac{a_{n-1} + a_{n-2}}{2} = \frac{(n-2)/(n-1) + 1}{2} = \frac{2n - 1}{2(n - 1)} $ $\ldots$
&
\textbf{Qwen3-30B-A3B-S\textsuperscript{3}} (13,098 tokens)\par Moreover, \textcolor{teal}{\textbf{the only way to get a\textsubscript{n}=1 is if all the terms being averaged (the last k terms) are equal to 1, because if any term in the average is less than 1, the average would be less than 1.}} \\
\midrule
\multicolumn{1}{p{0.47\linewidth}}{\textbf{Answer:} $\boxed{\dfrac{1}{4034}}$ $\times$} & \multicolumn{1}{p{0.47\linewidth}}{\textbf{Answer:} $\boxed{So the maximal possible value is\ldots}$ \textcolor{teal}{$\checkmark$}} \\
\bottomrule
\end{tabular}
\end{table}
\clearpage
\begin{table}[!htbp]
\centering
\small
\caption{\textbf{HMMT25, Example 8: A Polynomial Coefficient by Interpolation} (18.8\% token reduction). $S^3$ exploits $P'(x) \equiv -1$ in the interpolation; the baselines replace the coefficient with a count of the non-zero values.}
\vspace{6pt}
\label{tab:case9}
\begin{tabular}{p{0.47\linewidth} p{0.47\linewidth}}
\toprule
\multicolumn{2}{p{0.94\linewidth}}{\textbf{Problem:} Let $f$ be the unique polynomial of degree at most $2026$ such that for all $n \in\{1,2,3, \ldots, 2027\}$, $f(n)= \begin{cases}1 & \text { if } n \text { is a perfect square } \\ 0 & \text { otherwise }\end{cases}$ Suppose that $\frac{a}{b}$ is the coefficient of $x^{2025}$ in $f$, where $a$ and $b$ are integers such that $\operatorname{gcd}(a, b)=1$. Compute the unique integer $r$ between $0$ and $2026$ (inclusive) such that $a-r b$ is divisible by $2027$. (Note that $2027$ is prime.)} \\
\multicolumn{2}{p{0.94\linewidth}}{\textbf{Ground Truth:} 1037} \\
\midrule
\textbf{Qwen3-4B-Thinking} (18,715 tokens)\par \textcolor{red}{\textit{The leading coefficient $ C $ of $ f(x) $ satisfies:}} $ C \equiv -m \mod p $ where $ m $ is the number of perfect squares in $ \{1, 2, \ldots, p\} $ \textcolor{red}{\textit{This result is verified through multiple test cases with small primes and confirmed via detailed analysis}} $\ldots$
&
\textbf{Qwen3-4B-TIES} (22,348 tokens)\par \textcolor{red}{\textit{So $f(x)$ is degree at most 44?}}  \textcolor{red}{\textit{Wait, hold on! This is a critical point.}} The problem says "degree at most 2026", but if we have only 45 points where it's 1 and others 0, \textcolor{red}{\textit{isn't the interpolating polynomial of degree at most 44?}} $\ldots$ \\
\midrule
\multicolumn{1}{p{0.47\linewidth}}{\textbf{Answer:} 1982 $\times$} & \multicolumn{1}{p{0.47\linewidth}}{\textbf{Answer:} 0 $\times$} \\
\midrule
\textbf{Qwen3-4B-MI-0.8} (13,991 tokens)\par Therefore: \textcolor{red}{\textit{$ c_{2025} \equiv -45 \mod 2027. $}}  \textcolor{red}{\textit{To get the representative in $ [0, 2026] $, add 2027:}}  \textcolor{red}{\textit{$ r = (-45) \mod 2027 = 2027 - 45 = 1982. $}} --- Final Answer $\ldots$
&
\textbf{Qwen3-4B-S\textsuperscript{3}} (11,361 tokens)\par Differentiating both sides: \textcolor{teal}{\textbf{$ P'(x) = 2027 x^{2026} - 1 \equiv -1 \mod 2027 $}}  \textcolor{teal}{\textbf{So $ P'(n) = -1 $ for all $ n \in \mathbb{F}_{2027} $.}} \\
\midrule
\multicolumn{1}{p{0.47\linewidth}}{\textbf{Answer:} 1982 $\times$} & \multicolumn{1}{p{0.47\linewidth}}{\textbf{Answer:} 1037 \textcolor{teal}{$\checkmark$}} \\
\bottomrule
\end{tabular}
\end{table}
\clearpage
\begin{table}[!htbp]
\centering
\small
\caption{\textbf{Olympiad-Bench, Example 9: Sum of Squares over the Rationals} (53.6\% token reduction). $S^3$ reduces the question to how many rational squares 7 needs and answers with Legendre's theorem; Thinking and TIES try to exhibit a construction instead.}
\vspace{6pt}
\label{tab:case2}
\begin{tabular}{p{0.47\linewidth} p{0.47\linewidth}}
\toprule
\multicolumn{2}{p{0.94\linewidth}}{\textbf{Problem:} Find the smallest number $n$ such that there exist polynomials $f_{1}, f_{2}, \ldots, f_{n}$ with rational coefficients satisfying $x^{2}+7=f_{1}(x)^{2}+f_{2}(x)^{2}+\cdots+f_{n}(x)^{2} .$} \\
\multicolumn{2}{p{0.94\linewidth}}{\textbf{Ground Truth:} 5} \\
\midrule
\textbf{Qwen3-30B-A3B-Thinking} (31,456 tokens)\par I think I have to trust the theory here. \textcolor{red}{\textit{The key points are: - n=1: impossible, as shown. - n=2: impossible, as shown (requires 7 to be sum of two rational squares, which it's not). - n=3: impossible, as shown (7 cannot be sum of three rational squares). - n=4: possible, because 7 can be written as sum of four rational squares, and the orthogonality condition is solvable over Q (local conditions satisfied, so global solution exists by Hasse-Minkowski).}} Therefore, the minimal n is 4.
&
\textbf{Qwen3-30B-A3B-TIES} (15,776 tokens)\par Yes! \textcolor{red}{\textit{Here's a known example: $ x^2 + 7 = \left( \frac{x}{2} + \frac{3}{2} \right)^2 + \left( \frac{x}{2} + \frac{3}{2} \right)^2 + \left( \frac{x}{2} - \frac{3}{2} \right)^2 + \left( \frac{x}{2} - \frac{1}{2} \right)^2. $ Let's check: Each term: - $ (\frac{x}{2} + \frac{3}{2})^2 = \frac{x^2}{4} + \frac{3x}{2} + \frac{9}{4} $, - Same for the second, - $ (\frac{x}{2} - \frac{3}{2})^2 = \frac{x^2}{4} - \frac{3x}{2} + \frac{9}{4} $, - $ (\frac{x}{2} - \frac{1}{2})^2 = \frac{x^2}{4} - \frac{x}{2} + \frac{1}{4} $. Sum: - $ x^2 $: $ 4 \times \frac{1}{4} = 1 $, - $ x $: $ \frac{3}{2} + \frac{3}{2} - \frac{3}{2} - \frac{1}{2} = (3 - 2)/2 = 1/2 $, - Constants: $ 2 \times \frac{9}{4} + \frac{9}{4} + \frac{1}{4} = \frac{18+9+1}{4} = 28/4 = 7 $. Still has $ x $ term. But if we swap signs in the last two: Try: - $ f_1 = \frac{x}{2} + \frac{3}{2} $, - $ f_2 = \frac{x}{2} + \frac{3}{2} $, - $ f_3 = -\frac{x}{2} + \frac{3}{2} $, - $ f_4 = -\frac{x}{2} + \frac{1}{2} $. Then $ x $ terms: $ \frac{3}{2} + \frac{3}{2} + \frac{3}{2} + \frac{1}{2} = 4 $.}} Worse. $\ldots$ \\
\midrule
\multicolumn{1}{p{0.47\linewidth}}{\textbf{Answer:} 4 $\times$} & \multicolumn{1}{p{0.47\linewidth}}{\textbf{Answer:} 4 $\times$} \\
\midrule
\textbf{Qwen3-30B-A3B-MI-0.8} (32,175 tokens)\par From the second equation: $ ab = -cd \Rightarrow b = -\frac{c}{a}d $ (assuming $ a \ne 0 $). \textcolor{red}{\textit{Substitute into the third equation: $ b^2 + d^2 = \left(\frac{c^2}{a^2} + 1\right)d^2 = \frac{a^2 + c^2}{a^2} d^2 = \frac{1}{a^2} d^2. $ So: $ \frac{1}{a^2} d^2 + e^2 + f^2 = 7. $ Let $ u = \frac{d}{a} $, then $ u^2 + e^2 + f^2 = 7 $. This implies that 7 must be written as a sum of three rational squares, which is impossible (as shown earlier). Therefore, this configuration fails.}} Trying other configurations (e.g., three linear, one constant) similarly leads to contradictions or irrational coefficients $\ldots$
&
\textbf{Qwen3-30B-A3B-S\textsuperscript{3}} (14,931 tokens)\par But Legendre's three-square theorem says a natural number n is expressible as sum of three integer squares iff n is not of the form 4\textsuperscript{a}(8b + 7). \textcolor{teal}{\textbf{7 is exactly of the form 8b + 7 (b=0), so 7 cannot be written as sum of three integer squares.}} But can it be written as sum of four integer squares? \textcolor{teal}{\textbf{Yes, by Lagrange's four-square theorem, every natural number is sum of four integer squares.}} In fact, 7 = 2$^2$ + 1$^2$ + 1$^2$ + 1$^2$ = 4 + 1 + 1 + 1 = 7. \\
\midrule
\multicolumn{1}{p{0.47\linewidth}}{\textbf{Answer:} 5 $\checkmark$} & \multicolumn{1}{p{0.47\linewidth}}{\textbf{Answer:} 5 \textcolor{teal}{$\checkmark$}} \\
\bottomrule
\end{tabular}
\end{table}
\clearpage
\subsection{Vision Language Reasoning}
These six examples all turn on a figure --- a food web, a clock face, a
composite scene, a geometric diagram, and two accounting tables --- and in four of the six
no baseline answers correctly at all; in the other two exactly one does. The three analysed
below are Table~\ref{tab:case11}, where the hour hand is read as pointing at the wrong numeral;
Table~\ref{tab:case13}, where a spatial relation is never resolved; and Table~\ref{tab:case14},
where the baselines lose track of which quantity the table leaves blank.
\vspace{4pt}

\textbf{Representative examples.}

\textbf{Example 1 (Table~\ref{tab:case11}): Reading an Analogue Clock} (81.5\% reduction, incorrect $\rightarrow$ correct). A clock face, four candidate times, and one thing to read: where the two hands
point. The hour hand sits between the 4 and the 5, and the minute hand on the 6, so the time is
4:30 and the answer is (C) --- which is what $S^3$ reports, in 394 tokens. The three
baselines all report 5:30, and all three arrive there the same way: they read the hour hand as
pointing \emph{at} the 5 rather than between the 4 and the 5.

The traces are worth quoting because the error is visible in the first sentence and never
corrected. MI-0.8 (4{,}041) opens \textit{``It's pointing between 4 and 5. Wait, no,
actually, the hour hand is pointing towards the 5. Wait, no, wait.''} --- the correct reading is
produced and immediately discarded, and it is not recovered. TIES (2{,}130) reaches \textit{``So
answer D. But wait, maybe the hour hand is pointing to 5, which is 5:00, but the minute hand is
at 6, so 30 minutes. So 5:30. Yes, that makes sense. So the answer is D.''} and then re-derives
the same conclusion twice more. Thinking (8{,}073) resolves the
ambiguity the other way --- \textit{``The hour hand is positioned between 5 and 6''}, which the
image does not support --- and closes by explaining why a hand at 5:30 should sit half way
between the 5 and the 6: a rule stated correctly and applied to the wrong pair of numerals.

$S^3$ reads the hands once and keeps what it read: \textit{``the hour hand is between 4 and
5, the time must be 4:30''}. It then double-checks the one thing that could sink the reading ---
\textit{``sometimes people get confused if the hour hand is exactly on a number. But here, the
hour hand is just past 4''} --- and stops. The 81.5\% is not a shorter inspection; all four
traces inspect the same pixels. It is the cost of the other three reopening a reading.

\textbf{Example 2 (Table~\ref{tab:case13}): Tandem Bikes and Matte Trucks} (62.7\% reduction, incorrect $\rightarrow$ correct). A comparative question about a cluttered scene: are there fewer tandem bikes
behind the brown metal bicycle than matte trucks to the left of the green object? The answer is
no, because both counts are 1 --- the purple tandem bike is behind the brown bicycle, and the
gray pickup is the matte truck to the left of the green truck. $S^3$ counts 1 and 1 and
reports (B) in 551 tokens. The three baselines all count 0 and 1, all report (A) Yes, and all
three make the same mistake in the same place.

That place is a single word. MI-0.8 (2{,}124) decides \textit{``The purple tandem bike is to
the right of the brown bicycle, so it's in front of the brown bicycle''} --- it reads a
left/right position off the image and converts it directly into a front/behind relation, which
the image does not license. Thinking (7{,}059) reaches the same count through a longer route
that also fixes the wrong colour for the brown bicycle before correcting itself, and TIES
(1{,}478) arrives there through a chain of hedges: \textit{``Wait, no. The large gray truck is in
the center, and the green truck is behind it? Wait, no. The green truck is a small one, so the
large gray truck is to its left.''} In all three traces the second count is right and the first
is zero, so the comparison $0 < 1$ then decides the answer.

$S^3$ resolves the one relation the question depends on and then does the comparison:
\textit{``Behind it, there's a tandem bike''}, followed by \textit{``number of tandem bikes
behind brown bicycle: 1''}. It still has to decide whether a gray pickup counts as matte and
whether the bus counts as a truck, and it gets both of those right too --- but those two
decisions are the same in all four traces. The 62.7\% is the cost of the front/behind relation
not being settled.

\textbf{Example 3 (Table~\ref{tab:case14}): A Missing Figure in a Trial Balance} (75.6\% reduction, incorrect $\rightarrow$ correct). An accounting table lists several companies, one cell of company D's row is
blank, and four candidate amounts are offered. The blank cell is company D's \emph{expenses};
company D's revenues, $1{,}054{,}116$, are printed in the table. So the problem is one
substitution into $\text{net income} = \text{revenues} - \text{expenses} + \text{gains} -
\text{losses}$, and $S^3$ performs exactly that, in 1{,}463 tokens, to get
$1{,}023{,}309 + 58{,}275 = 1{,}081{,}584$ and option (C). All three baselines write down the
same equation and then lose track of which of its terms is the one being asked for.

Thinking (10{,}382 tokens) works the equation for each option in turn, computing the expenses
that would correspond to each candidate revenue, and finishes by choosing on magnitude:
\textit{``The options are all around 1 million, so 1,210,732 is the highest option. Wait, maybe
the correct answer is D.''} TIES (6{,}003) computes the same four candidate expenses and then
asks itself \textit{``Now, is there any other information?''} before settling on A. MI-0.8
(8{,}611) is the most explicit about the failure --- it cannot decide whether the missing cell is
expenses or revenues, says so, and then answers anyway: \textit{``maybe the question is to find
Expenses, but the options are Revenues. No. Wait, maybe the question is to find the missing
amount for Company D, which is Expenses, and the correct answer is A.''} Three traces, three
different wrong options, and in each the arithmetic is sound.

$S^3$ spends its first line identifying which cell is blank and then never revisits the
question: it reads company D's revenues off the table, substitutes once, and matches the result
to the option list. The 75.6\% is the difference between one substitution and four.
\begin{table}[!htbp]
\centering
\small
\caption{\textbf{MathVista-testmini, Example 10: Ferns and Salmon in a Food Web} (87.1\% token reduction). $S^3$ follows the one edge the question turns on; the baselines reconstruct the whole food web from memory.}
\vspace{6pt}
\label{tab:case10}
\begin{tabular}{p{0.47\linewidth} p{0.47\linewidth}}
\toprule
\multicolumn{2}{@{}p{0.94\linewidth}@{}}{\begin{minipage}[c]{0.465\linewidth}\centering\setlength{\fboxsep}{2pt}\setlength{\fboxrule}{0.6pt}\fbox{\includegraphics[width=4.8cm]{figs/case_study/mathvista_370.jpg}}\end{minipage}\hspace{0.02\linewidth}\begin{minipage}[c]{0.455\linewidth}\textbf{Problem:} Hint: Please answer the question and provide the correct option letter, e.g., A, B, C, D, at the end. Question: In the diagram of the food web shown, if the number of ferns decrease, the supply of salmon will most likely? Choices: (A) decrease (B) can't tell (C) stay same (D) increase\par\vspace{4pt}\textbf{Ground Truth:} decrease\end{minipage}} \\
\midrule
\textbf{Qwen3-VL-4B-Thinking} (5,741 tokens)\par \textcolor{red}{\textit{Wait, maybe the user made a mistake in the diagram description, but assuming the diagram is correct, we have to go with the connections given.}}  \textcolor{red}{\textit{Since the diagram shows Ferns connected to Rabbits, and Salmon connected to Insects: Ants (connected to Flowers), there's no direct or indirect connection between Ferns and Salmon.}} Therefore, the supply of Salmon would stay the same. So the answer is C) stay same.
&
\textbf{Qwen3-VL-4B-TIES} (2,167 tokens)\par \textcolor{red}{\textit{Therefore, we can't tell.}}  \textcolor{red}{\textit{Wait, but let's check again.}} The diagram has arrows from Ferns to Small Mammals: Rabbits. So rabbits eat ferns. Then, what do rabbits eat? If ferns are the main food, then fewer ferns $\to$ fewer rabbits. \textcolor{red}{\textit{Now, do rabbits eat salmon?}} $\ldots$ \\
\midrule
\multicolumn{1}{p{0.47\linewidth}}{\textbf{Answer:} C $\times$} & \multicolumn{1}{p{0.47\linewidth}}{\textbf{Answer:} B $\times$} \\
\midrule
\textbf{Qwen3-VL-4B-MI-0.8} (2,705 tokens)\par \textcolor{red}{\textit{Is there an arrow from Ferns to Salmon?}}  \textcolor{red}{\textit{Let me think.}} If the number of ferns decreases, what happens to salmon? \textcolor{red}{\textit{If ferns are a food source for salmon, then less ferns would mean less food for salmon, so salmon population would decrease.}}
&
\textbf{Qwen3-VL-4B-S\textsuperscript{3}} (280 tokens)\par \textcolor{teal}{\textbf{Wait, in the diagram, there's a line from ferns to salmon.}} So ferns are a food source for salmon. If ferns decrease, that means there's less food available for salmon. \textcolor{teal}{\textbf{So salmon's supply would likely decrease because they have less food.}} \\
\midrule
\multicolumn{1}{p{0.47\linewidth}}{\textbf{Answer:} C $\times$} & \multicolumn{1}{p{0.47\linewidth}}{\textbf{Answer:} A \textcolor{teal}{$\checkmark$}} \\
\bottomrule
\end{tabular}
\end{table}
\clearpage
\begin{table}[!htbp]
\centering
\small
\caption{\textbf{MathVista-testmini, Example 11: Reading an Analogue Clock} (81.5\% token reduction). $S^3$ reads both hands once; the baselines re-open the hour hand a dozen times.}
\vspace{6pt}
\label{tab:case11}
\begin{tabular}{p{0.47\linewidth} p{0.47\linewidth}}
\toprule
\multicolumn{2}{@{}p{0.94\linewidth}@{}}{\begin{minipage}[c]{0.465\linewidth}\centering\setlength{\fboxsep}{2pt}\setlength{\fboxrule}{0.6pt}\fbox{\includegraphics[width=2.8cm]{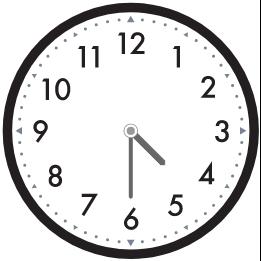}}\end{minipage}\hspace{0.02\linewidth}\begin{minipage}[c]{0.455\linewidth}\textbf{Problem:} Hint: Please answer the question and provide the correct option letter, e.g., A, B, C, D, at the end. Question: What time does the clock show? Choices: (A) 9:30 (B) 1:30 (C) 4:30 (D) 5:30 (E) 11:30\par\vspace{4pt}\textbf{Ground Truth:} 4:30\end{minipage}} \\
\midrule
\textbf{Qwen3-VL-4B-Thinking} (8,073 tokens)\par \textcolor{red}{\textit{Wait, this is really hard without seeing the image.}}  \textcolor{red}{\textit{Wait, but the user provided a description: "the clock show" with numbers 1-12.}}  \textcolor{red}{\textit{The hour hand is pointing to 5. Or wait, no.}} $\ldots$
&
\textbf{Qwen3-VL-4B-TIES} (2,130 tokens)\par \textcolor{red}{\textit{This is a bit tricky.}}  \textcolor{red}{\textit{Wait, perhaps the image shows the hour hand pointing to 5.}} But if the minute hand is at 6 (30 minutes), then the hour hand should be halfway between the current hour and the next. So if it's 5:30, $\ldots$ \\
\midrule
\multicolumn{1}{p{0.47\linewidth}}{\textbf{Answer:} D $\times$} & \multicolumn{1}{p{0.47\linewidth}}{\textbf{Answer:} D $\times$} \\
\midrule
\textbf{Qwen3-VL-4B-MI-0.8} (4,041 tokens)\par \textcolor{red}{\textit{Wait, the user provided a clock image. Let me try to imagine it.}}  \textcolor{red}{\textit{The clock has numbers 1 through 12.}}  \textcolor{red}{\textit{The minute hand is pointing at the 6 (which is 30 minutes), so the time is X:30.}}
&
\textbf{Qwen3-VL-4B-S\textsuperscript{3}} (394 tokens)\par Looking at the image, \textcolor{teal}{\textbf{the hour hand is pointing just past the 4, and the minute hand is pointing at the 6.}} \\
\midrule
\multicolumn{1}{p{0.47\linewidth}}{\textbf{Answer:} D $\times$} & \multicolumn{1}{p{0.47\linewidth}}{\textbf{Answer:} C \textcolor{teal}{$\checkmark$}} \\
\bottomrule
\end{tabular}
\end{table}
\clearpage
\begin{table}[!htbp]
\centering
\small
\caption{\textbf{MathVista-testmini, Example 12: Which Angle Is Numbered Two} (67.3\% token reduction). $S^3$ takes the two labels at $A$ as the two parts of the right angle and subtracts once; Thinking and MI argue over the vertex order in ``ABDC'' and set angle 2 equal to angle 1.}
\vspace{6pt}
\label{tab:case12}
\begin{tabular}{p{0.47\linewidth} p{0.47\linewidth}}
\toprule
\multicolumn{2}{@{}p{0.94\linewidth}@{}}{\begin{minipage}[c]{0.465\linewidth}\centering\setlength{\fboxsep}{2pt}\setlength{\fboxrule}{0.6pt}\fbox{\includegraphics[width=5.2cm]{figs/case_study/mathvista_499.png}}\end{minipage}\hspace{0.02\linewidth}\begin{minipage}[c]{0.455\linewidth}\textbf{Problem:} Hint: Please answer the question and provide the correct option letter, e.g., A, B, C, D, at the end. Question: Quadrilateral $ABDC$ is a rectangle. If $m\angle1 = 38$, find $m \angle 2$ Choices: (A) 33 (B) 38 (C) 52 (D) 87\par\vspace{4pt}\textbf{Ground Truth:} 52\end{minipage}} \\
\midrule
\textbf{Qwen3-VL-2B-Thinking} (16,634 tokens)\par The problem says quadrilateral ABDC is a rectangle, so ABDC is a rectangle, which means AB is parallel to CD, and AD is parallel to BC. Also, all angles in a rectangle are right angles, so angle at B is 90 degrees, right? \textcolor{red}{\textit{Wait, the diagram shows a rectangle with points A, B, D, C, so the vertices are A, B, D, C, making ABDC a rectangle. So AB is one side, BD is a diagonal, and CD is the opposite side. Wait, no, in a rectangle, the order of the vertices matters.}} $\ldots$
&
\textbf{Qwen3-VL-2B-TIES} (7,072 tokens)\par First, let's recall what we know about rectangles. A rectangle has all right angles, so angle at B is 90 degrees. Also, in a rectangle, opposite sides are equal, and diagonals are equal and bisect each other. \textcolor{red}{\textit{Wait, but here we have quadrilateral ABDC as a rectangle. Wait, let's check the diagram. First, let's identify the rectangle. The diagram shows a rectangle with vertices A, B, D, C. Wait, the vertices are A, B, D, C. So ABDC is a rectangle.}} $\ldots$ \\
\midrule
\multicolumn{1}{p{0.47\linewidth}}{\textbf{Answer:} B $\times$} & \multicolumn{1}{p{0.47\linewidth}}{\textbf{Answer:} C $\checkmark$} \\
\midrule
\textbf{Qwen3-VL-2B-MI-0.8} (18,735 tokens)\par \textcolor{red}{\textit{Wait, hold on, the vertices are A, B, D, C? Wait, the diagram shows a rectangle with points A, B, C, D. Let me check the labels. The rectangle is ABDC? Wait, no, the rectangle is probably ABCD, but the problem says ABDC. Wait, maybe it's a typo, but let's look at the diagram. The diagram has points A, B, C, D with a right angle at B, so AB is a side, BC is a side, CD is the other side, DA is the last side. Wait, the right angle is at B, so angle ABC is 90 degrees. That makes sense for a rectangle: all angles are 90 degrees, opposite sides equal.}} $\ldots$
&
\textbf{Qwen3-VL-2B-S\textsuperscript{3}} (2,315 tokens)\par \textcolor{teal}{\textbf{Now consider $\angle 2$, which is at vertex $A$, between side $AB$ and diagonal $AD$. Since $AB$ and $AD$ are adjacent sides of the rectangle, $\angle BAD = 90^\circ$. We already have: $ \angle BAC = 38^\circ $ So: $ \angle CAD = \angle BAD - \angle BAC = 90^\circ - 38^\circ = 52^\circ $ Thus: $ \angle 2 = \angle CAD = 52^\circ $}} $\ldots$ \\
\midrule
\multicolumn{1}{p{0.47\linewidth}}{\textbf{Answer:} B $\times$} & \multicolumn{1}{p{0.47\linewidth}}{\textbf{Answer:} C \textcolor{teal}{$\checkmark$}} \\
\bottomrule
\end{tabular}
\end{table}
\clearpage
\begin{table}[!htbp]
\centering
\small
\caption{\textbf{MathVista-testmini, Example 13: Tandem Bikes and Matte Trucks} (62.7\% token reduction). $S^3$ resolves ``behind'' once and compares; the baselines never settle the spatial relation.}
\vspace{6pt}
\label{tab:case13}
\begin{tabular}{p{0.47\linewidth} p{0.47\linewidth}}
\toprule
\multicolumn{2}{@{}p{0.94\linewidth}@{}}{\begin{minipage}[c]{0.465\linewidth}\centering\setlength{\fboxsep}{2pt}\setlength{\fboxrule}{0.6pt}\fbox{\includegraphics[width=4.4cm]{figs/case_study/mathvista_502.jpg}}\end{minipage}\hspace{0.02\linewidth}\begin{minipage}[c]{0.455\linewidth}\textbf{Problem:} Hint: Please answer the question and provide the correct option letter, e.g., A, B, C, D, at the end. Question: Are there fewer tandem bikes that are behind the brown metal bicycle than matte trucks on the left side of the green object? Choices: (A) Yes (B) No\par\vspace{4pt}\textbf{Ground Truth:} No\end{minipage}} \\
\midrule
\textbf{Qwen3-VL-2B-Thinking} (7,059 tokens)\par \textcolor{red}{\textit{Let's list all objects:}}  \textcolor{red}{\textit{1. Small green bus (left side, gray and green) 2. Gray road bike (left of the truck) 3. Blue airplane (middle)}}  \textcolor{red}{\textit{4. Small blue and green truck (green object? Wait, the green object is the truck with green color.}} Wait, the green truck is the one with the green color, so the truck is green. Then, the other truck is the gray pickup truck in the center. $\ldots$
&
\textbf{Qwen3-VL-2B-TIES} (1,478 tokens)\par \textcolor{red}{\textit{Wait, "behind" in 3D space---relative to the brown bike.}}  \textcolor{red}{\textit{The brown bike is a road bike, and the tandem bike is a little to the right and maybe behind?}}  \textcolor{red}{\textit{Wait, no. Let's visualize.}} $\ldots$ \\
\midrule
\multicolumn{1}{p{0.47\linewidth}}{\textbf{Answer:} A $\times$} & \multicolumn{1}{p{0.47\linewidth}}{\textbf{Answer:} A $\times$} \\
\midrule
\textbf{Qwen3-VL-2B-MI-0.8} (2,124 tokens)\par \textcolor{red}{\textit{so behind the brown bicycle? Wait, no---position matters.}} The brown bicycle (road bike) is on the right side. \textcolor{red}{\textit{The tandem bike is to the right of the brown bicycle, so maybe behind?}}  \textcolor{red}{\textit{Wait, no. Let's look at the scene.}} $\ldots$
&
\textbf{Qwen3-VL-2B-S\textsuperscript{3}} (551 tokens)\par Let's see: \textcolor{teal}{\textbf{the tandem bike is purple and located behind the brown bicycle.}}  \textcolor{teal}{\textbf{So that's one tandem bike.}} \\
\midrule
\multicolumn{1}{p{0.47\linewidth}}{\textbf{Answer:} A $\times$} & \multicolumn{1}{p{0.47\linewidth}}{\textbf{Answer:} B \textcolor{teal}{$\checkmark$}} \\
\bottomrule
\end{tabular}
\end{table}
\clearpage
\begin{table}[!htbp]
\centering
\small
\caption{\textbf{MMMU, Example 14: A Missing Figure in a Trial Balance} (75.6\% token reduction). $S^3$ identifies which cell is missing; the baselines solve for a cell that was already given.}
\vspace{6pt}
\label{tab:case14}
\begin{tabular}{p{0.47\linewidth} p{0.47\linewidth}}
\toprule
\multicolumn{2}{@{}p{0.94\linewidth}@{}}{\begin{minipage}[c]{0.465\linewidth}\centering\setlength{\fboxsep}{2pt}\setlength{\fboxrule}{0.6pt}\fbox{\includegraphics[width=5.6cm]{figs/case_study/mmmu_test_Accounting_34.png}}\end{minipage}\hspace{0.02\linewidth}\begin{minipage}[c]{0.455\linewidth}\textbf{Problem:} Each of the following situations relates to a different company. [Image 1] For company D, find the missing amounts. A. $1,034,325 B.$1,098,650 C. $1,081,584 D.$1,210,732 Answer with the option letter only.\par\vspace{4pt}\textbf{Ground Truth:} C\end{minipage}} \\
\midrule
\textbf{Qwen3-VL-4B-Thinking} (10,382 tokens)\par \textcolor{red}{\textit{Wait, maybe the question is asking for Revenues?}}  \textcolor{red}{\textit{Let's see.}}  \textcolor{red}{\textit{Wait, let's rearrange the formula for Revenues: Revenues = Net Income + Expenses - Gains + Losses}} Wait, let's see. From Net Income = Revenues - Expenses + $\ldots$
&
\textbf{Qwen3-VL-4B-TIES} (6,003 tokens)\par \textcolor{red}{\textit{Since all options satisfy the equation, but the problem must have one correct answer, maybe I made a mistake in the problem}} $\ldots$ \\
\midrule
\multicolumn{1}{p{0.47\linewidth}}{\textbf{Answer:} D $\times$} & \multicolumn{1}{p{0.47\linewidth}}{\textbf{Answer:} A $\times$} \\
\midrule
\textbf{Qwen3-VL-4B-MI-0.8} (8,611 tokens)\par \textcolor{red}{\textit{So all options work. But that can't be.}}  \textcolor{red}{\textit{There must be a mistake in the problem or my reasoning.}}  \textcolor{red}{\textit{Wait, maybe the question is to find the missing amount for Company D, which is Expenses, and the}} $\ldots$
&
\textbf{Qwen3-VL-4B-S\textsuperscript{3}} (1,463 tokens)\par \textcolor{teal}{\textbf{Wait, so the only missing amount for Company D is Expenses.}}  \textcolor{teal}{\textbf{Let's verify.}} Net Income = Revenues - Expenses + Gains - Losses \textcolor{teal}{\textbf{We can rearrange to solve for Expenses: Expenses = Revenues + Gains - Losses}} \\
\midrule
\multicolumn{1}{p{0.47\linewidth}}{\textbf{Answer:} A $\times$} & \multicolumn{1}{p{0.47\linewidth}}{\textbf{Answer:} C \textcolor{teal}{$\checkmark$}} \\
\bottomrule
\end{tabular}
\end{table}
\clearpage
\begin{table}[!htbp]
\centering
\small
\caption{\textbf{MMMU, Example 15: Shifts in a Loanable-Funds Diagram} (95.5\% token reduction). $S^3$ moves one curve per shock and reads the new crossing off the diagram; MI sends the fall in capital spending the wrong way and TIES moves the rise in saving onto the demand curve.}
\vspace{6pt}
\label{tab:case15}
\begin{tabular}{p{0.47\linewidth} p{0.47\linewidth}}
\toprule
\multicolumn{2}{@{}p{0.94\linewidth}@{}}{\begin{minipage}[c]{0.465\linewidth}\centering\setlength{\fboxsep}{2pt}\setlength{\fboxrule}{0.6pt}\fbox{\includegraphics[width=5.8cm]{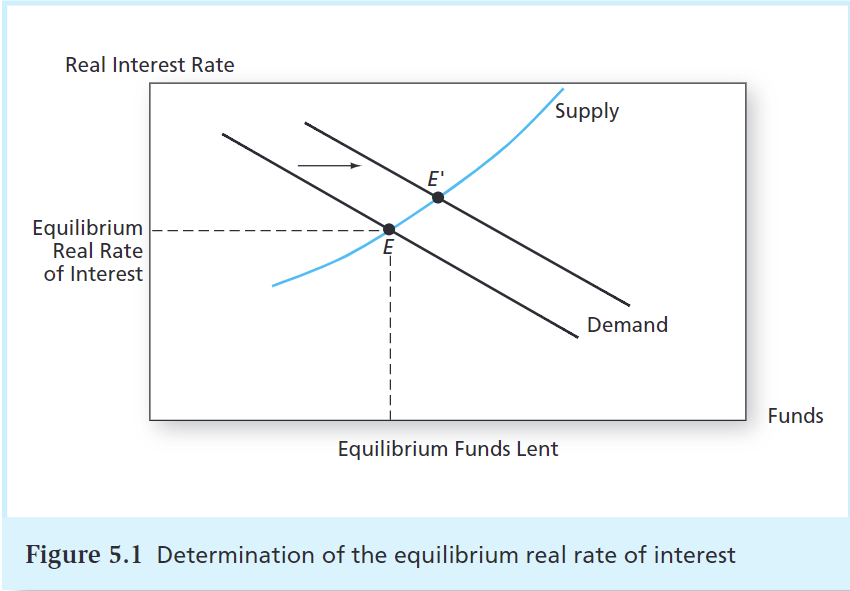}}\end{minipage}\hspace{0.02\linewidth}\begin{minipage}[c]{0.455\linewidth}\textbf{Problem:} Use the image to analyze the effect of the following on the level of real interest rates:[Image 1] a. Businesses become more pessimistic about future demand for their products and decide to reduce their capital spending. b. Households are induced to save more because of increased uncertainty about their future Social Security benefits. c. The Federal Reserve Board undertakes open-market purchases of U.S. Treasury securities in order to increase the supply of money. A. a.increase ;b.decrease ;c.increase B. a.increase ;b.decrease ;c.decrease C. a.decrease ;b.decrease ;c.decrease D. a.decrease ;b.increase ;c.decrease Answer with the option letter only.\par\vspace{4pt}\textbf{Ground Truth:} C\end{minipage}} \\
\midrule
\textbf{Qwen3-VL-2B-Thinking} (7,620 tokens)\par So for option a: demand for funds (for investment) decreases $\to$ demand curve shifts left $\to$ intersection is to the left of E $\to$ interest rate is lower (decrease). \textcolor{red}{\textit{Option b: households save more $\to$ demand for funds (for investment) decreases $\to$ demand curve shifts left $\to$ interest rate is lower (decrease).}} Option c: Federal Reserve buys Treasury securities $\to$ supply of money increases $\to$ supply curve shifts right $\to$ intersection is to the right of E $\to$ interest rate is lower (decrease). So all three options decrease the interest rate. $\ldots$
&
\textbf{Qwen3-VL-2B-TIES} (919 tokens)\par \textcolor{red}{\textit{D. a.decrease ;b.increase ;c.decrease $\to$ yes, that's D.}} Wait, let me confirm again. For a: businesses reduce capital spending $\to$ demand for funds decreases $\to$ demand curve shifts left $\to$ equilibrium real rate decreases $\to$ correct. \textcolor{red}{\textit{For b: households save more $\to$ demand for funds increases $\to$ demand curve shifts right $\to$ equilibrium real rate increases $\to$ correct.}} For c: Fed buys Treasury securities $\to$ increases money supply $\to$ supply of funds shifts right $\to$ equilibrium real rate decreases $\to$ correct. So the answer is D. $\ldots$ \\
\midrule
\multicolumn{1}{p{0.47\linewidth}}{\textbf{Answer:} C $\checkmark$} & \multicolumn{1}{p{0.47\linewidth}}{\textbf{Answer:} D $\times$} \\
\midrule
\textbf{Qwen3-VL-2B-MI-0.8} (1,023 tokens)\par In the supply and demand model for funds, the demand curve is the demand for funds by businesses, households, etc. \textcolor{red}{\textit{If businesses reduce their capital spending, that's a decrease in the demand for funds (since they're not borrowing as much to spend on capital). So the demand curve shifts left. The intersection of supply and demand shifts left, so the equilibrium real rate of interest increases. Wait, but in the graph, the supply curve is upward sloping, and demand is downward sloping.}} $\ldots$
&
\textbf{Qwen3-VL-2B-S\textsuperscript{3}} (341 tokens)\par So a is decrease. b. Households save more due to uncertainty about Social Security benefits. \textcolor{teal}{\textbf{Saving more means more funds in the banking system, so the supply of funds increases. Supply of funds shifts right.}} That would lead to a lower real interest rate. Because with more supply, the equilibrium interest rate falls. So b is decrease. c. Federal Reserve buys Treasury securities, increasing money supply. \textcolor{teal}{\textbf{That shifts the supply curve for funds to the right (since more money is in the economy). So supply increases, equilibrium interest rate decreases.}} So c is decrease. $\ldots$ \\
\midrule
\multicolumn{1}{p{0.47\linewidth}}{\textbf{Answer:} B $\times$} & \multicolumn{1}{p{0.47\linewidth}}{\textbf{Answer:} C \textcolor{teal}{$\checkmark$}} \\
\bottomrule
\end{tabular}
\end{table}
\clearpage
\subsection{Audio Reasoning}
These five examples are audio questions: one listening-comprehension item,
three from the MMAR set of audio-reasoning questions, and one disfluency-annotation item from
MMSU. In four of the five no baseline answers correctly; the fifth, Table~\ref{tab:case18}, is
a clip on which all four methods succeed, and it is included for contrast. Each of the other
four turns on a single cue, and each of the three analysed below turns on a different kind:
Table~\ref{tab:case17}, where the culprit is identified by what is said rather than by how
someone reacts; Table~\ref{tab:case19}, where the baselines discount a demonstration that has
already been given; and Table~\ref{tab:case20}, where one opening sound has to be classified
before anything else can be counted.
\vspace{4pt}

\textbf{Representative examples.}

\textbf{Example 1 (Table~\ref{tab:case17}): Who Ate the Powdered Donut} (55.0\% reduction, incorrect $\rightarrow$ correct). A four-speaker exchange, and one question: who ate the donut? The fourth
speaker says so --- \textit{``Yeah, that was me.''} --- and $S^3$ quotes exactly that,
naming the fourth person in 218 tokens. All three baselines hear the same exchange, all three
reconstruct the culprit from something other than the admission, and all three land on the wrong
speaker.

Thinking (3{,}929 tokens) and MI-0.8 (3{,}006) build their answer on a cough. MI-0.8
reasons \textit{``The physical reaction (coughing) aligns with the act of consuming something
like a powdered donut, and Speaker 3's denial followed by this sound indicates guilt''};
Thinking lays the same inference out as a four-row elimination and concludes \textit{``the third
speaker (Speaker 3) ate the powdered donut''}. The cough is real, and the inference from it is
not unreasonable --- but it is a guess about a speaker who never says anything, made in a clip
where another speaker does. TIES (484) takes the third route, from ownership to guilt:
\textit{``The speaker who then makes the statement about it being `his last donut' and calls
the others `disrespectful' is also the first speaker''}, and answers A.

$S^3$ transcribes the same four turns and reads the fourth one for what it is, an admission:
\textit{``Speaker 4 (fourth person) admits by saying, ``Yeah, that was me.'',\,''} It then confirms
the fourth person is option D and stops. The 55.0\% is the cost of the two competing inferences that
$S^3$ does not enter.

\textbf{Example 2 (Table~\ref{tab:case19}): Counting the Chinese Tones Demonstrated} (70.9\% reduction, incorrect $\rightarrow$ correct). The speaker teaches four Mandarin tones, demonstrates each one individually,
and then says six syllables that use them. The question asks how many different tones are
\emph{involved in the six pronunciation words}, and the answer key says four --- which requires
counting the individual demonstration, since the six words on their own use three. $S^3$
takes that reading and answers (D) in 535 tokens. All three baselines take the other reading and
answer (A) 3.

What makes this example useful is that the baselines find the missing fourth tone and then
discard it. MI-0.8 (5{,}912) states the whole case and then rules it out in one sentence:
\textit{``Although the speaker mentioned learning four tones earlier in the audio with the
``a'' sounds, the specific six words provided do not include the fourth tone.''} TIES
(1{,}841) and
Thinking (4{,}147) reach the same conclusion without ever considering the demonstration, both
listing the same three tones and both ending on the same negative finding --- Thinking's is
\textit{``The fourth tone (à) is not demonstrated in any of the six syllables.''} Each of the
three then reports the count of the narrower set.

$S^3$ keeps both parts of the clip in scope: it records the four individual demonstrations
and then the six words, and finds the fourth tone among the former. The 70.9\% is the cost of
the six-syllable inventory the baselines build.

\textbf{Example 3 (Table~\ref{tab:case20}): Types of Disfluency in the Audio} (41.8\% reduction, incorrect $\rightarrow$ correct). The clip opens with a short laugh, then a woman describes a trifle: \textit{``Ha,
well my sister is the simstrous of the two of us.''} The options ask which disfluency categories
are present, and the answer is A, discourse markers and restarts --- \textit{well} for the
first, \textit{simstrous} for the second. $S^3$ answers A in 271 tokens. All three baselines
answer D, and all three do so because they classify the opening laugh as a filled pause.

The pair of traces to compare is $S^3$'s and MI-0.8's, because they transcribe the same
sound differently. MI-0.8 (1{,}944) hears \textit{``Huh''} and files it with \textit{um} and
\textit{uh}; TIES (1{,}459) does the same; Thinking (466) calls it \textit{``a short, sharp
laugh''} and then files it as a filled pause anyway --- \textit{``This is a common filled pause,
often used to indicate''}. $S^3$ makes the same observation and draws the opposite
conclusion: \textit{``The initial ``Ha'' is a chuckle, which is a filler but not a standard filled
pause like `uh' or `um'.''} The rest of the classification follows. MI-0.8 also shows the
baselines overriding their own taxonomy --- \textit{``So even though discourse markers aren't
disfluencies, the test might be considering them as such''} --- to keep the option that the
misclassification produced.

$S^3$ then checks the remaining options off and stops. The 41.8\% is the smallest reduction
of the twenty, and the difference between the two traces is a single perceptual decision taken
in the first line of both.
\begin{table}[!htbp]
\centering
\small
\caption{\textbf{MMAR, Example 16: A Profession Inferred from a Conversation} (42.4\% token reduction). $S^3$ takes the word ``test'' as the domain cue; the baselines supply a fitness practice that does not exist.}
\vspace{6pt}
\label{tab:case16}
\begin{tabular}{p{0.47\linewidth} p{0.47\linewidth}}
\toprule
\multicolumn{2}{p{0.94\linewidth}}{\textbf{Problem:} What is the most likely profession of the first speaker?} \\
\multicolumn{2}{p{0.94\linewidth}}{\textbf{Audio transcript} \textit{(reader aid; the model received only the audio)}: ``all right bro looking good we're done yeah we're all done man what's up there you go this is for you you want me to eat this yeah man candy yeah sure all right that was a test dude which you failed''} \\
\multicolumn{2}{p{0.94\linewidth}}{\textbf{Ground Truth:} Dentist} \\
\midrule
\textbf{Qwen3-Omni-30B-A3B-Thinking} (1,978 tokens)\par The key part is that they're talking about candy, and then one says it was a test that the other failed. But what does this have to do with professions? Now, the question is: What is the most likely profession of the first speaker? Options are Dentist, Teacher, Fitness Coach, Nurse. I need to think about contexts where someone might give candy as a test or something related to health. Let me consider each option: \textcolor{red}{\textit{- Dentist: Dentists often talk about oral health, and candy might be related to teeth. But dentists usually advise against eating candy because it's bad for teeth.}} However, in this case, the first speaker gives candy and says it was a test. \textcolor{red}{\textit{Maybe it's a test for something like how people react to candy, but that seems odd.}}
&
\textbf{Qwen3-Omni-30B-A3B-TIES} (549 tokens)\par \textcolor{red}{\textit{The first speaker says, 'All right, bro. Looking good. We're done?' The second speaker asks, 'We're done?' and the first speaker confirms, 'Yeah, we're all done, man.' The second speaker then says, 'What's up?' and the first speaker responds, 'There you go. This is for you.' The second speaker asks, 'You want me to eat this?' and the first speaker says, 'Yeah, man. Why not? The candy.}} $\ldots$ \\
\midrule
\multicolumn{1}{p{0.47\linewidth}}{\textbf{Answer:} C $\times$} & \multicolumn{1}{p{0.47\linewidth}}{\textbf{Answer:} C $\times$} \\
\midrule
\textbf{Qwen3-Omni-30B-A3B-MI-0.8} (7,446 tokens)\par \textcolor{red}{\textit{Perhaps for a dentist, after a check-up, they might say "Looking good," and then give a small treat, but candy is bad for teeth, so it doesn't make sense for a test.}} Unless it's to see if the patient resists, but the patient is given candy to eat, so if they eat it, they fail, but dentists don't usually do that. For a teacher, it's unlikely. For a nurse, similar. Let's think about the word "test." \textcolor{red}{\textit{In medical contexts, there are glucose tests, but candy isn't used for that.}}
&
\textbf{Qwen3-Omni-30B-A3B-S\textsuperscript{3}} (316 tokens)\par \textcolor{teal}{\textbf{Crucially, the word 'test' is used in what appears to be a professional context, suggesting a formal evaluation.}} \\
\midrule
\multicolumn{1}{p{0.47\linewidth}}{\textbf{Answer:} C $\times$} & \multicolumn{1}{p{0.47\linewidth}}{\textbf{Answer:} A \textcolor{teal}{$\checkmark$}} \\
\bottomrule
\end{tabular}
\end{table}
\clearpage
\begin{table}[!htbp]
\centering
\small
\caption{\textbf{MMAR, Example 17: Who Ate the Powdered Donut} (55.0\% token reduction). $S^3$ reads the fourth speaker's admission; the baselines infer the culprit from a cough.}
\vspace{6pt}
\label{tab:case17}
\begin{tabular}{p{0.47\linewidth} p{0.47\linewidth}}
\toprule
\multicolumn{2}{p{0.94\linewidth}}{\textbf{Problem:} Which speaker ate the powdered donut} \\
\multicolumn{2}{p{0.94\linewidth}}{\textbf{Audio transcript} \textit{(reader aid; the model received only the audio)}: ``Hey,\allowbreak{} wait my last powder donut it wasn't me wasn't me Guys that was his last powder donut. Don't be so disrespectful. Okay?''} \\
\multicolumn{2}{p{0.94\linewidth}}{\textbf{Ground Truth:} Fourth person} \\
\midrule
\textbf{Qwen3-Omni-30B-A3B-Thinking} (3,929 tokens)\par Here, "his" refers to Speaker 1 (the owner), confirming the donut was eaten, and the phrase "Don't be so disrespectful" implies Speaker 4 is addressing the eater. \textcolor{red}{\textit{The coughing directly after Speaker 3's denial suggests they are the one who ate the donut, as the physical reaction gives away their guilt.}} Speaker 1 is the owner and not the eater, Speaker 2 denied responsibility without coughing, and Speaker 4 is accusing others, so they are not the eater. Therefore, the third speaker (Speaker 3) ate the powdered donut.
&
\textbf{Qwen3-Omni-30B-A3B-TIES} (484 tokens)\par Speaker 2 (female) says, 'It wasn't me,' which is a direct denial of eating it. Speaker 1 then says, 'Wasn't me,' which is another denial. \textcolor{red}{\textit{After these denials, Speaker 1 starts coughing, and then says, 'Guys, that was his last powdered donut. Don't be so disrespectful. Okay, fresh up right now.'}} This is the key part. $\ldots$ \\
\midrule
\multicolumn{1}{p{0.47\linewidth}}{\textbf{Answer:} B $\times$} & \multicolumn{1}{p{0.47\linewidth}}{\textbf{Answer:} A $\times$} \\
\midrule
\textbf{Qwen3-Omni-30B-A3B-MI-0.8} (3,006 tokens)\par The audio: "Wasn't me." then "cough cough cough Oh!" then "Guys, that was his last powdered donut..." So coughing is separate from Speaker 4's speech. Therefore, the coughing is from one of the previous speakers. \textcolor{red}{\textit{Since Speaker 3 just said "Wasn't me," and then coughs, it's likely Speaker 3.}}  \textcolor{red}{\textit{Speaker 2 spoke earlier, so less likely. Therefore, Speaker 3 ate it.}}
&
\textbf{Qwen3-Omni-30B-A3B-S\textsuperscript{3}} (218 tokens)\par Then, analyzing the audio reveals four speakers: Speaker 1 (first person) asks, "Hey, when the last powdered donut?", Speaker 2 (second person) denies, "It wasn't me.", Speaker 3 (third person) denies, "Wasn't me.", \textcolor{teal}{\textbf{and Speaker 4 (fourth person) admits by saying, "Yeah, that was me.", followed by coughing and gagging, then clarifying, "That was his last powdered donut. Don't be so disrespectful."}}  \textcolor{teal}{\textbf{This admission occurs when the speaker identifies themselves as the one who ate it.}} \\
\midrule
\multicolumn{1}{p{0.47\linewidth}}{\textbf{Answer:} B $\times$} & \multicolumn{1}{p{0.47\linewidth}}{\textbf{Answer:} D \textcolor{teal}{$\checkmark$}} \\
\bottomrule
\end{tabular}
\end{table}
\clearpage
\begin{table}[!htbp]
\centering
\small
\caption{\textbf{MMAU\_testmini, Example 18: The Ingredients of a Trifle} (59.6\% token reduction). $S^3$ recalls the layers once and dismisses each wrong list in a clause; the baselines re-check every distractor ingredient by ingredient.}
\vspace{6pt}
\label{tab:case18}
\begin{tabular}{p{0.47\linewidth} p{0.47\linewidth}}
\toprule
\multicolumn{2}{p{0.94\linewidth}}{\textbf{Problem:} What ingredients did Speaker 1 say she used in the trifle?} \\
\multicolumn{2}{p{0.94\linewidth}}{\textbf{Audio transcript} \textit{(reader aid; the model received only the audio)}: ``it's a trifle it's got all of these layers it's got all of these layers first there's a layer of ladyfingers then a layer of jam then custard which i made from scratch then raspberries more ladyfingers then beef sauteed with peas and onions and then bananas and then i just put some whipped cream on top it tastes like feet i like it are you kidding''} \\
\multicolumn{2}{p{0.94\linewidth}}{\textbf{Ground Truth:} (D) Ladyfingers, jam, custard, raspberries, beef, peas, onions, bananas, whipped cream} \\
\midrule
\textbf{Qwen3-Omni-30B-A3B-Thinking} (1,373 tokens)\par A. Bread, jelly, pudding, strawberries, chicken, carrots, garlic, apples, cream \textcolor{red}{\textit{- Bread? Not mentioned; she said lady fingers, which are like cookies or biscuits, but not bread. - Jelly? She said jam, which is similar, but option says jelly. - Pudding? She said custard, which might be confused, but custard is different. - Strawberries? She said raspberries. - Chicken? She said beef.}} - Carrots, garlic, apples? Not mentioned. - Cream? She said whipped cream, but option says cream, which might be close. $\ldots$
&
\textbf{Qwen3-Omni-30B-A3B-TIES} (589 tokens)\par Option A includes bread, jelly, pudding, strawberries, chicken, carrots, garlic, apples, cream. \textcolor{red}{\textit{'Bread' is not correct as 'ladyfingers' were mentioned; 'jelly' is not correct as 'jam' was used; 'strawberries' is incorrect as 'raspberries' were stated; 'chicken' is wrong because 'beef' was used; and other items like 'carrots' and 'garlic' were not mentioned, so A is incorrect.}} $\ldots$ \\
\midrule
\multicolumn{1}{p{0.47\linewidth}}{\textbf{Answer:} D $\checkmark$} & \multicolumn{1}{p{0.47\linewidth}}{\textbf{Answer:} D $\checkmark$} \\
\midrule
\textbf{Qwen3-Omni-30B-A3B-MI-0.8} (1,728 tokens)\par Option A: Bread, jelly, pudding, strawberries, chicken, carrots, garlic, apples, cream \textcolor{red}{\textit{- Bread: she said lady fingers, which are a type of cookie or biscuit, not bread. But lady fingers are often used in trifles, so maybe similar, but not exactly bread. - Jelly: she said jam, which is similar, but jelly is usually fruit spread, jam is thicker. But she said jam. - Pudding: she said custard, which is different. Pudding is usually set with gelatin or cornstarch, custard is egg-based. Not the same. - Strawberries: she said raspberries, not strawberries.}} $\ldots$
&
\textbf{Qwen3-Omni-30B-A3B-S\textsuperscript{3}} (238 tokens)\par \textcolor{teal}{\textbf{Next, I compared these ingredients with each option: Option A includes bread, strawberries, chicken, carrots, and garlic which were not mentioned; Option B includes cake, marmalade, cherries, turkey, corn, peppers, and melons which were not mentioned; Option C includes cookies, blueberries, pork, beans, tomatoes, peaches, and yogurt which were not mentioned; Option D lists ladyfingers, jam, custard, raspberries, beef, peas, onions, bananas, and whipped cream which exactly matches Speaker 1's description. Therefore, the correct answer is D.}} $\ldots$ \\
\midrule
\multicolumn{1}{p{0.47\linewidth}}{\textbf{Answer:} D $\checkmark$} & \multicolumn{1}{p{0.47\linewidth}}{\textbf{Answer:} D \textcolor{teal}{$\checkmark$}} \\
\bottomrule
\end{tabular}
\end{table}
\clearpage
\begin{table}[!htbp]
\centering
\small
\caption{\textbf{MMAR, Example 19: Counting the Chinese Tones Demonstrated} (70.9\% token reduction). $S^3$ transcribes the last syllable as ``m\'a''; the baselines hear ``m\u{a}'' and lose the fourth tone.}
\vspace{6pt}
\label{tab:case19}
\begin{tabular}{p{0.47\linewidth} p{0.47\linewidth}}
\toprule
\multicolumn{2}{p{0.94\linewidth}}{\textbf{Problem:} How many different Chinese tones are involved in the six Chinese pronunciation words demonstrated by the speaker?} \\
\multicolumn{2}{p{0.94\linewidth}}{\textbf{Audio transcript} \textit{(reader aid; the model received only the audio)}: ``Chinese is easy when you learn four tones in this way. Move your head with the tone shape. Try this.''} \\
\multicolumn{2}{p{0.94\linewidth}}{\textbf{Ground Truth:} 4} \\
\midrule
\textbf{Qwen3-Omni-30B-A3B-Thinking} (4,147 tokens)\par Each syllable's tone is as follows: - First "m$\bar{a}$" in "mama": first tone ($\bar{a}$) - Second "m$\bar{a}$" in "mama": first tone ($\bar{a}$) - "q$\acute{\imath}$": second tone ($\acute{a}$) - "m$\check{a}$" in "q$\acute{\imath}$ m$\check{a}$": third tone ($\check{a}$) \textcolor{red}{\textit{- "m$\check{a}$" in "m$\check{a}$ m$\bar{a}$": third tone ($\check{a}$) - "m$\bar{a}$" in "m$\check{a}$ m$\bar{a}$": first tone ($\bar{a}$)}} The tones used are first, second, and third. \textcolor{red}{\textit{The fourth tone ($\grave{a}$) is not demonstrated in any of the six syllables.}} Therefore, there are t $\ldots$
&
\textbf{Qwen3-Omni-30B-A3B-TIES} (1,841 tokens)\par \textcolor{red}{\textit{3. m$\check{a}$ m$\bar{a}$}} - "m$\check{a}$" is third tone. \textcolor{red}{\textit{- "m$\bar{a}$" is first tone.}} Summary of Tones Used: - First tone: appears in "m$\bar{a}$" (repeatedly). - Second tone: appears in "q$\acute{\imath}$". - Third tone: appears in "m$\check{a}$" (repeatedly). \textcolor{red}{\textit{The fourth tone is not present in any of the six syllables.}} \\
\midrule
\multicolumn{1}{p{0.47\linewidth}}{\textbf{Answer:} A $\times$} & \multicolumn{1}{p{0.47\linewidth}}{\textbf{Answer:} A $\times$} \\
\midrule
\textbf{Qwen3-Omni-30B-A3B-MI-0.8} (5,912 tokens)\par \textcolor{red}{\textit{She says: "$\bar{a}$, $\acute{a}$, $\check{a}$, $\grave{a}$" -- that's the four tones. Then she says "Try this: m$\bar{a}$, m$\bar{a}$, q$\acute{\imath}$, m$\check{a}$, m$\check{a}$, m$\check{a}$" -- let's parse that. She says: "m$\bar{a}$, m$\bar{a}$, q$\acute{\imath}$, m$\check{a}$, m$\check{a}$, m$\check{a}$" -- but that might be a typo or mispronunciation. Looking back: "m$\bar{a}$, m$\bar{a}$, q$\acute{\imath}$, m$\check{a}$, m$\check{a}$, m$\check{a}$" -- but "m$\bar{a}$" is repeated, and "q$\acute{\imath}$" is different. Let me read the audio transcript: "Try this: m$\bar{a}$, m$\bar{a}$, q$\acute{\imath}$, m$\check{a}$, m$\check{a}$, m$\check{a}$" -- but it says "m$\bar{a}$ m$\bar{a}$ q$\acute{\imath}$ m$\check{a}$ m$\check{a}$ m$\check{a}$" -- that seems odd. Perhaps it's "m$\bar{a}$, m$\bar{a}$, q$\acute{\imath}$, m$\check{a}$, m$\check{a}$, m$\check{a}$" meaning six words. But "m$\bar{a}$" is first tone, "q$\acute{\imath}$" is second tone, "m$\check{a}$" is third tone, and "m$\grave{a}$" might be fourth tone, but she says "m$\check{a}$" multiple times. Let's look at the audio: she says "m$\bar{a}$, m$\bar{a}$, q$\acute{\imath}$, m$\check{a}$, m$\check{a}$, m$\check{a}$" -- but "m$\check{a}$" is third tone, and she says it three times? That doesn't make sense for different tones.}}
&
\textbf{Qwen3-Omni-30B-A3B-S\textsuperscript{3}} (535 tokens)\par Then in the examples: \textcolor{teal}{\textbf{- m$\bar{a}$ m$\bar{a}$: both tone 1 - q$\acute{\imath}$ m$\check{a}$: q$\acute{\imath}$ is tone 2, m$\check{a}$ is tone 3 - m$\check{a}$ m$\acute{a}$: m$\check{a}$ is tone 3, m$\acute{a}$ is tone 4}}  \textcolor{teal}{\textbf{So the tones used are tone 1, tone 2, tone 3, tone 4.}}  \textcolor{teal}{\textbf{All four tones are demonstrated.}} \\
\midrule
\multicolumn{1}{p{0.47\linewidth}}{\textbf{Answer:} A $\times$} & \multicolumn{1}{p{0.47\linewidth}}{\textbf{Answer:} D \textcolor{teal}{$\checkmark$}} \\
\bottomrule
\end{tabular}
\end{table}
\clearpage
\begin{table}[!htbp]
\centering
\small
\caption{\textbf{MMSU, Example 20: Types of Disfluency in Speech} (41.8\% token reduction). $S^3$ hears the opening sound as a chuckle; the baselines count it as a filled pause.}
\vspace{6pt}
\label{tab:case20}
\begin{tabular}{p{0.47\linewidth} p{0.47\linewidth}}
\toprule
\multicolumn{2}{p{0.94\linewidth}}{\textbf{Problem:} Which types of disfluencies are present in the audio? - Filled pauses: e.g., uh, um - Discourse markers: e.g., well, you know - Restarts: interrupted or repeated sentence starts - Explicit editing terms: e.g., I mean - None: if the speech is fluent.} \\
\multicolumn{2}{p{0.94\linewidth}}{\textbf{Audio transcript} \textit{(reader aid; the model received only the audio)}: ``Well,\allowbreak{} my sister is the seamstress of the two of us,\allowbreak{} but she doesn't China paint,\allowbreak{} so we both have our expertise side.''} \\
\multicolumn{2}{p{0.94\linewidth}}{\textbf{Ground Truth:} discourse markers, restarts} \\
\midrule
\textbf{Qwen3-Omni-30B-A3B-Thinking} (466 tokens)\par \textcolor{red}{\textit{Then, the audio starts with a short, sharp laugh: "Heh." This is a common filled pause, often used to indicate a brief hesitation or as a conversational filler. Next, the speaker says: "Well my sister is the seamstress of the two of us..." The word "Well" is a classic discourse marker used to introduce a statement.}}
&
\textbf{Qwen3-Omni-30B-A3B-TIES} (1,459 tokens)\par \textcolor{red}{\textit{First, listen to the speech: "Huh, well my sister is the sinstrous of the two of us. And but she doesn't chain up paint, so we both have our expertise site." Now, break down each type: 1. Filled pauses: The speaker says "Huh" at the beginning. "Huh" is a filled pause, similar to "um" or "uh", used to fill a moment of hesitation.}} \\
\midrule
\multicolumn{1}{p{0.47\linewidth}}{\textbf{Answer:} D $\times$} & \multicolumn{1}{p{0.47\linewidth}}{\textbf{Answer:} D $\times$} \\
\midrule
\textbf{Qwen3-Omni-30B-A3B-MI-0.8} (1,944 tokens)\par \textcolor{red}{\textit{Then, analyzing the audio content: the speaker says, 'Huh. Well, my sister is the seamstress of the two of us, and but she doesn't China paint, so we both have our expertise side.' Specifically, at the very beginning, there's a 'Huh' which is a filled pause (a short, non-lexical sound expressing hesitation or thought), and around 00:03, the speaker says 'and but' which is a classic example of a restart}} $\ldots$
&
\textbf{Qwen3-Omni-30B-A3B-S\textsuperscript{3}} (271 tokens)\par \textcolor{teal}{\textbf{First, listen to the beginning: "Ha, well my sister is the simstrous of the two of us..." - The initial "Ha" is a chuckle, which is a filler but not a standard filled pause like "uh" or "um". - The word "well" is a discourse marker, used to introduce the sentence. - The phrase "simstrous" is a self-correction for "simstress". This is a restart/repetition of a word.}} \\
\midrule
\multicolumn{1}{p{0.47\linewidth}}{\textbf{Answer:} D $\times$} & \multicolumn{1}{p{0.47\linewidth}}{\textbf{Answer:} A \textcolor{teal}{$\checkmark$}} \\
\bottomrule
\end{tabular}
\end{table}

\end{document}